\documentclass[review]{elsarticle}

\usepackage{lineno,hyperref}
\usepackage{subcaption}
\usepackage{booktabs,array}
\usepackage{url}
\usepackage{booktabs}
\usepackage[usenames, dvipsnames]{xcolor}
\usepackage{amsmath} 
\usepackage{amssymb}  
\usepackage{multicol}
\usepackage{algorithm, algpseudocode}
\definecolor{myGrey}{rgb}{0.9,0.9,0.9}
\newcommand{\norm}[1]{\left\lVert#1\right\rVert}

\modulolinenumbers[5]

\journal{Journal of Robotics and Autonomous Systems}

\definecolor{thegreen}{rgb}{0.06,0.55,0.08}
\begin{document}
\begin{frontmatter}

\title{Learning Context-Adaptive Task Constraints for Robotic Manipulation}

\author[mymainaddress]{Dennis Mronga\corref{mycorrespondingauthor}}
\author[mymainaddress,mysecondaryaddress]{Frank Kirchner}

\cortext[mycorrespondingauthor]{Corresponding author}
\address[mymainaddress]{Robotics Innovation Center of German Research Center for Artificial Intelligence GmbH (DFKI), Bremen, Germany}
\address[mysecondaryaddress]{Robotics Group of the University of Bremen, Bremen, Germany}

\begin{abstract}

Constraint-based control approaches offer a flexible way to specify robotic manipulation tasks and execute them on robots with many degrees of freedom. However, the specification of task constraints and their associated priorities usually requires a human-expert and often leads to tailor-made solutions for specific situations. This paper presents our recent efforts to automatically derive task constraints for a constraint-based robot controller from data and adapt them with respect to previously unseen situations (contexts). We use a programming-by-demonstration approach to generate training data in multiple variations (context changes) of a given task. From this data we learn a probabilistic model that maps context variables to task constraints and their respective soft task priorities. We evaluate our approach with 3 different dual-arm manipulation tasks on an industrial robot and show that it performs better than comparable approaches with respect to reproduction accuracy in previously unseen contexts. 

\end{abstract}

\begin{keyword}
Context-Adaptive Control, Constraint-Based Robot Control, Programming-by-Demonstration, Gaussian Mixture Regression, Dual-Arm Manipulation
\end{keyword}

\end{frontmatter}


\section{INTRODUCTION} \label{sec:introduction}

Many robotic manipulation tasks like bi-manual handling of an object, polishing a table or opening a door can be described as a combination of simpler tasks. For example, the problem of polishing a table can be decomposed into "maintain surface contact" and "follow trajectory". Apart from that robotic manipulation tasks are usually subject to constraints, which may be related to the environment (e.g., properties of the contacted surface), to restrictions of the given task (e.g., a container with liquid that must not be tilted) or physical limitations of the robot (e.g., joint limits).  

Constraint-based control, also referred to as task-oriented or Whole-Body Control, offers a flexible way to deal with such (constrained) multi-task problems. It formulates simultaneously running tasks as constraints to an instantaneous optimization problem, where the computed optimum represents the robot joint command that best accomplishes all the tasks. This way, multiple robot tasks can be integrated, complex control problems can be composed from simpler (sub-)tasks and the degrees of freedom (dof) of the entire robot body can be exploited. Within the last years a large number of frameworks have been proposed that allow multi-task control on velocity~\cite{Smits2008}, acceleration~\cite{Flacco2012} or torque~\cite{Dietrich2012} level. Most of these frameworks use of some kind of prioritization strategy in order to facilitate the parallel execution of possibly conflicting tasks. Depending on the type of prioritization, the selected task priorities are referred to as either strict~\cite{Sentis2006} or soft~\cite{Dehio2015}, while some frameworks also allow a mixture of both types~\cite{Liu2016}.

Even though constraint-based control is a proven tool to specify complex control problems, it requires a human expert to model the overall problem in terms of task constraints and associated priorities. This process is mostly performed manually, which is time-consuming, error-prone and leads to solutions, which are often tailored to a specific situation. If the specification of the given task or the environment changes, these handcrafted solutions will likely fail. 

In order to overcome these issues we develop an approach to (a) automatically derive task constraints for robotic manipulation and their associated soft priorities from data and (b) generalize about task variations and adapt to previously unseen situations. The data is obtained by the means of a programming-by-demonstration approach and the tasks are varied in between the demonstrations. 

Throughout this paper, we refer to these task variations as \textit{context changes}. Generally spoken, \textit{context} in robotics can be defined as "a configuration of features which are (...) useful to influence the decision process of a robotic system"~\cite{Bloisi2016}. Approaches that are able to automatically adapt the robot controls with respect to such changes are referred to as \textit{context-adaptive}. As an adaptation model, we use a Dirichlet Process Gaussian Mixture Model (DPGMM~\cite{Neal1992}), which models the joint distribution of \textit{context variables} and task constraints. Using this probabilistic model, we use Gaussian Mixture Regression (GMR)~\cite{Calinon2007} for reproduction of the task constraints and their associated priorities. 
 
This paper is organized as follows: Section~\ref{sec:related_work} presents a summary of the related work on automatic derivation and generalization of constraints in task-oriented control frameworks. Section~\ref{sec:control_framework} gives a quick overview on the constraint-based control framework. In Section~\ref{sec:learning_task_constraints} we illustrate our methods on learning adaptive task constraints from demonstration. In Section~\ref{sec:results} we show experimental results and discuss possible extensions and future works in Section~\ref{sec:conclusion}.

Throughout the document we use the notations and symbols shown in Table~\ref{tab:symbols}. Vectors are represented by lowercase bold characters, matrices by uppercase bold characters.

\section{RELATED WORK} \label{sec:related_work}

Constraint-based control is a powerful tool to program robots with many degrees of freedom and it has been applied to increasingly complex robotic tasks throughout the years. However, nearly all the available approaches leave the task specification to the skilled programmer, which has to model motion and physical constraints of the robot, select task priorities and tune task parameters in a cumbersome, mostly manual trial-and-error procedure. Even worse, the resulting task specification usually performs well only in a limited context. If the task or environment changes, the task parameters have to be adapted again. In our work we want to provide a way for the non-expert to program complex robotic systems using constraint-based control. To achieve this, we use programming-by-demonstration methods to record data from robotic manipulation tasks and derive task constraints, as well as their associated task priorities from this data using probabilistic regression models. By demonstrating the tasks in varying contexts the models are able to adapt the reproduced task constraints with respect to a variety of context changes that the task is subject to. 

Different works exist that also attempt to ease the burden of the human programmer and automatize the process of selecting task constraints and/or priorities for constraint-based frameworks. A number of approaches apply constrained stochastic optimization or reinforcement learning to find task priorities that improve the overall robot behavior e.g., in terms of robustness~\cite{Charbonneau2018}, safety~\cite{Modugno2016a}, constraint satisfaction~\cite{Modugno2016, Lober2016}, smoothness of motion~\cite{Mronga2020} or generalization capabilities~\cite{Dehio2015}. Compared to our work these approaches focus on the automatic derivation of (soft) task priorities in terms of mixing weights that balance the contribution of different (predefined) task constraints. Here, we want derive both, task constraints and their respective priorities from data. Furthermore, the aforementioned methods provide only limited generalization capabilities by optimizing task priorities with respect to one particular situation. Our approach on the other hand attempts to generalize task constraints over a variety of situations. 

In robot behavior learning, a widespread approach is to learn initial trajectories by imitation and refine them using reinforcement learning, where the behaviors are often represented by a movement model, for example dynamic movement primitives (DMP)~\cite{Kober2010}. DMP's themselves have been designed to generalize over some meta-parameters like initial position or movement duration. The capability to adapt to more complex context changes can be achieved by the means of hierarchical approaches, where an upper-level policy is learned that generalizes over the meta-parameters of the lower-level policy~\cite{Fabisch2014, Kupcsik2017, Wilbers2017}. However, these methods typically focus on a single task that is executed on a robot with six or seven dof, while we on the other hand focus on multi-task scenarios on more complex systems. 

Learning task constraints from user-demonstrations has also been dealt with before. For example Armesto et al.~\cite{Armesto2017} learn wiping a smooth surface with a 7-dof arm  by separately estimating a task policy and a nullspace constraint that generalizes over previously unseen contexts (in this case the orientation of the surface). They use fixed task priorities and a fixed hierarchy. Compared to that, in our work we additionally want to estimate the task priorities from the demonstrated motions. The work of Perico et al.~\cite{Perico2019} combines a constraint-based control framework with imitation learning. They represent a demonstrated trajectory using a probabilistic model and integrate it as a constraint in their control framework. The variance of the demonstrated trajectories is thereby used to modulate the stiffness of the robot and guide the human operator towards the estimated target. The other task parameters (e.g., the task priorities) are still selected manually. In contrast to that, we want to use the variability of the user-demonstrations to obtain an estimate of all the task priorities. Also, the generalization capabilities are limited to variations of the target position and orientation of the end effector, whereas we want to generalize over more complex task parameters. In~\cite{Silverio2017} the authors extend the probabilistic movement model developed in~\cite{Calinon2016} to additionally learn task priorities from demonstration. For that they use a soft weighting scheme for a manually selected set of candidate hierarchies. In contrast, our approach relies on soft task priorities and does not require the selection of candidate hierarchies. In~\cite{Fang2016} Random Forest Regression is combined with constraint-based robot control in order to learn a pouring task. The required training data is generated by naive user-demonstrations in an interactive simulated environment. However, this method is somewhat specific to the problem of pouring liquid into a container, while we attempt to provide a more general approach. 

Yet another promising research direction is to parametrize constraint-based controllers by the means of high-level reasoning mechanisms~\cite{Leidner2016, Tenorth2014}. However, here the task parameters to reason about still have to be selected manually or at least some range of allowed values has to be provided by the human expert. In a sense these approaches do not automatize the task specification process, but  shift the problem of parameter selection on a higher, more user-friendly level. 

\begin{table}
\newcolumntype{P}[1]{>{\centering\arraybackslash}p{#1}}
\newcolumntype{L}[1]{>{\raggedright\arraybackslash}p{#1}}
\newcolumntype{R}[1]{>{\raggedleft\arraybackslash}p{#1}}
 \centering
 \footnotesize
 \def\arraystretch{1.2}
 \begin{tabular}{lP{1.5cm}P{1.6cm}P{1.5cm}P{1.5cm}P{1.5cm}P{1.5cm}}
  \midrule
   Approach & Use PbD & Task Constraints & Task Priorities &  Soft/Strict Priorities & Genera-lization  \\
   \midrule 
   \cite{Modugno2016a},\cite{Modugno2016} & - & - &  $\color{thegreen}\surd$ & soft & - \\
   \cite{Lober2016} & - & $\color{thegreen}\surd$ & - & soft & - \\
   \cite{Dehio2015} & - & - & $\color{thegreen}\surd$ & soft & - \\
   \cite{Armesto2017} & $\color{thegreen}\surd$ & $\color{thegreen}\surd$ & - & strict & $\color{thegreen}\surd$ \\
   \cite{Perico2019} & $\color{thegreen}\surd$ & $\color{thegreen}\surd$ & - & soft & $\color{thegreen}\surd$\\
   \cite{Silverio2017} & $\color{thegreen}\surd$ & $\color{thegreen}\surd$ & $\color{thegreen}\surd$ & strict & $\color{thegreen}\surd$\\
   \cite{Fang2016} & $\color{thegreen}\surd$ & $\color{thegreen}\surd$ & - & soft & $\color{thegreen}\surd$ \\
  \midrule
   Our approach & $\color{thegreen}\surd$ & $\color{thegreen}\surd$ &  $\color{thegreen}\surd$ & soft & $\color{thegreen}\surd$\\
  \midrule
 \end{tabular}
 \caption{Comparison of our approach with the state of the art.}
 \label{tab:sota}
\end{table}

Table~\ref{tab:sota} compares our approach with similar state-of-the-art methods according to the type of learning (Programming-by-demonstration vs. other, e.g., Reinforcement Learning), the learning target (task constraints, task priorities or both), whether the approach uses strict or soft priorities and whether or not the approach achieves generalization capabilities. According to the table the most similar approach to ours is~\cite{Silverio2017}. However, this approach demands that the user selects candidate hierarchies in terms of projection operators in advance. Compared to that, our approach only requires the selection of task relevant coordinate frames, which is trivial in virtually every case. Furthermore the approach in ~\cite{Silverio2017} relies on strict task hierarchies, which does not work well on over-constrained problems as we will show in section~\ref{sec:results_prio_compare}. In summary, the main differences and innovations of our approach compared to existing state of the art methods are

\begin{itemize}
\item We simultaneously estimate task constraints and their respective (soft) task priorities, while most other approaches deal with either the one or the other
\item There is no need to manually select candidate constraints or task hierarchies with our approach.
\item Most existing approaches generalize over different starting or target positions. In contrast to that, we focus on major context changes, which may lead to completely different characteristics of a given task.
\end{itemize}

\begin{table}
\tiny
\noindent\fcolorbox{myGrey}{myGrey}{%
\begin{minipage}{0.48\textwidth}
\begin{tabular}{p{2cm}p{3cm}}
\multicolumn{2}{l}{\textbf{Operators}} \\
$\dot{x}$ & Time derivative of $x$\\
$\hat{x}$ & Estimate of $x$\\
${A}^{-1}$ & Inverse of a matrix $\mathbf{A}$\\
${A}^{T}$ & Transpose of matrix $\mathbf{A}$\\
$[\mathbf{x}]$ & Skew symmetric matrix of $\mathbf{x}$ (see e.g.,~\cite{lynch2017})\\
$\text{tr}(\mathbf{A})$ & Trace of a square matrix $\mathbf{A}$\vspace{0.3cm}\\
\multicolumn{2}{l}{\textbf{Dimensions}} \\
$N$ & Number of robot joints\\
$M$ & Number of task constraints\\
$D$ & Number of user demonstrations per context\\
$C$ & Number of context variables \\
$S$ & Number of samples per demonstration \\
$L$ & Number of constraint variables \\
$K$ & Number of mixture components \\
$F$ & Number of task frames\vspace{0.3cm} \\
\multicolumn{2}{l}{\textbf{Robot Control}} \\
$\mathbf{x} \in \mathbb{R}^{6}$ & Pose in Cartesian space  \\ 
$\mathbf{p} \in \mathbb{R}^{3}$ & Position in Cartesian space   \\ 
\end{tabular}
\end{minipage}
\hfill
\begin{minipage}{0.48\textwidth}
\begin{tabular}{p{2cm}p{3cm}}
$\mathbf{R} \in \mathrm{SO}(3)$ & Rotation matrix\\ 
$\mathbf{v} \in \mathbb{R}^{6}$  & Twist in Cartesian space \\
$\mathbf{K}\in \mathbb{R}^{6 \times 6}$ & Diagonal gain matrix \\ 
$\mathbf{A} \in \mathbb{R}^{6 \times N}$ & Task Jacobian  \\ 
$\theta\in \mathbb{R}$ & Rotation angle  \\ 
$\mathbf{\hat{\omega}} \in \mathbb{R}^{3}$ & Unit rotation axis\\ 
$\mathbf{q} \in \mathbb{R}^{N}$ & Robot joint positions \\ 
$\mathbf{W} \in \mathbb{R}^{6 \times 6}$ & Diagonal task weight matrix \\ 
$\mathbf{w} \in \mathbb{R}^{6}$ & Task weight vector \\ 
$\Delta t$ & Sample time \vspace{0.3cm}  \\ 
\multicolumn{2}{l}{\textbf{Mixture Models}} \\
$\mathcal{P}(x)$ & Probability distribution of $x$\\ 
$\mathbf{\mu}$ &  Mean of a Gaussian \\ 
$\mathbf{\Sigma}$ &  Covariance matrix of a Gaussian \\ 
${\sigma}^2$ &  Variance \\ 
$\pi$ & Mixing weight in a GMM \vspace{0.3cm} \\ 
\multicolumn{2}{l}{\textbf{Data Sets and modeling}} \\
$\mathbf{\kappa} \in \mathbb{R}^{C}$ & Context vector \\
$\mathbf{\xi} $ & Multi-dimensional data set \\
$\mathbf{X} \in \mathbb{R}^{D \cdot S \times L} $ & Data set with poses  \\
$\mathbf{V} \in \mathbb{R}^{D \cdot S \times L} $ & Data set with twists  \\
$\mathbf{\mathcal{K}} \in \mathbb{R}^{D \cdot S \times C}$ & Data set with contexts \\
\end{tabular}
\end{minipage}
}%
\caption{Overview of notations and variable names}
\label{tab:symbols}
\end{table}

\section{CONSTRAINT-BASED CONTROL FRAMEWORK} \label{sec:control_framework}

The control framework that we use is an adaptation of the approach in~\cite{Smits2008}. For controlling the pose of a robot link in Cartesian space, we use a proportional controller with feed forward term\footnote{In most equations we omit the dependence on time or robot joint state, for the sake of better readability.}

\begin{equation} \label{eq:pose_control}
\renewcommand*{\arraystretch}{1.2}
\mathbf{v}_d=\mathbf{v}_{r} + \mathbf{K}_p\left(\begin{matrix}\mathbf{p}_r-\mathbf{p}_a \\ \mathbf{R}_a\cdot \theta \boldsymbol{\hat \omega}^a_r\end{matrix} \right)
\end{equation}
where $\mathbf{v}_d \in \mathbb{R}^{6}$ is the twist that represents the control output composed of linear and angular velocity,  $\mathbf{v}_r \in \mathbb{R}^{6}$ is the desired (feed forward) twist and $\mathbf{K}_p \in  \mathbb{R}^{6 \times 6}$ is a diagonal matrix containing 6 feedback gain constants. The vectors $\mathbf{p}_r \in \mathbb{R}^{3}$ and $\mathbf{p}_a \in \mathbb{R}^{3}$ denote the reference and actual position of the controlled robot frame. The vector $\theta \boldsymbol{\hat \omega}_r^a \in \mathbb{R}^3$ comprises the exponential coordinates of the rotation matrix $\mathbf{R}^a_r=\mathbf{R}^{-1}_a\mathbf{R}_r \in SO(3)$, where $\mathbf{R}_a, \mathbf{R}_r \in SO(3)$ refer to the actual and reference orientation. This representation, which is closely related to the angle-axis representation of rotations, can be computed using the matrix logarithm of rotations~\cite{lynch2017}. It represents the rotation axis and angle, which rotates the frame $\mathbf{R}_a$, such that its orientation matches $\mathbf{R}_r$. This term is multiplied by $\mathbf{R}_a$ in order to transform to the base frame of the robot.

For each robot task, we define a controller according to Equation~(\ref{eq:pose_control}) and represent its control output as a constraint in the following online optimization problem\footnote{Here, only Cartesian position and orientation constraints are considered. However, the framework is also able to deal with other types of constraints like joint limits, collision avoidance or contact forces.} 

\begin{equation}
\begin{array}{ccc}\label{eq:opt_problem}
\renewcommand{\arraystretch}{1.5}
 \underset{\mathbf{\dot{q}}}{\text{minimize}} & \norm{\mathbf{\dot{q}}}_{2} & \\ 
 \text{subject to} & \left( \begin{matrix} \mathbf{A}_{w,1} \\ \vdots \\  \mathbf{A}_{w,M} \end{matrix} \right)  \mathbf{\dot{q}}
 =\left( \begin{matrix} \mathbf{v}_{d,1} \\ \vdots \\  \mathbf{v}_{d,M} \end{matrix} \right) \\
\end{array}
\end{equation}

where $\mathbf{\dot{q}} \in \mathbb{R}^{N}$ is the robot's reference joint velocity, $N$ the number of robot joints, $M$ is the number of task constraints and $\mathbf{A}_{w,i} = \mathbf{WA}_i \in \mathbb{R}^{6 \times N}$ is the weighted task Jacobian related to the $i$-th task.  The term $\mathbf{W} \in \mathbb{R}^{6 \times 6}$ is a diagonal matrix containing the \textit{task weights} $\mathbf{w}=(w_1\ldots w_6)$. The solution of Equation~(\ref{eq:opt_problem}) is computed using the damped Pseudo Inverse method as described in~\cite{Maciejewski1988}. The task weights thereby balance the importance of the constraint variables. For example, when controlling only the position of the robot in Cartesian space, the orientation might be irrelevant, so the corresponding task weights can be set to zero. This means the tasks are not hierarchically organized as in~\cite{Sentis2006}, but the solution is computed as a weighted combination of the control outputs. In the over-constrained case, an approximate solution will be assumed, governed by the values of the weights.

We prefer task weights, also referred to as \textit{soft task priorities}, over strict hierarchies here, since they facilitate the application of machine learning methods as described in the next section. In the following sections, we will use the terms "task weights" and "soft task priorities" equivalently. 

The term \textit{task constraint} is not uniquely defined in the robotics literature. Here, we define a task constraint as a pose/twist pair $(\mathbf{x}(t),\mathbf{v}(t))$ with associated soft task priorities $\mathbf{w}(t)$. Each of these pose/twist pairs describes the relative motion of two robot coordinate frames during task execution and can be used as input to the controller in Equation~(\ref{eq:pose_control}). Since the task constraints and soft task priorities are functions of time,  they may change during task execution and can be described as time-indexed trajectories. 

In the following section we will described an approach to automatically derive the quantities $(\mathbf{x}(t),\mathbf{v}(t),\mathbf{w}(t))$ from user demonstrations.

\section{LEARNING ADAPTIVE TASK CONSTRAINTS FROM DEMONSTRATION} \label{sec:learning_task_constraints}

The design of task constraints and selection of task weights as described in the previous section is usually done by an expert in a manual fashion. This process is time-consuming and the resulting motions are often tailored to a specific situation. An automated procedure that derives the reference input for the controller in Equation (\ref{eq:pose_control}) and the corresponding task weights used in Equation (\ref{eq:opt_problem}) could not only ease the burden of the programmer, but also lead to better results, especially if the solution can be adapted automatically to context changes. Such context changes could refer to the task itself (e.g., goal positions, orientation constraints, ...), the environment (e.g., size or shape of objects, position and moving direction of obstacles, ...) or the morphology of the robot (e.g., single arm or dual arm, with/without mobile base). 

\begin{figure}
\centering
\includegraphics[width=0.8\linewidth]{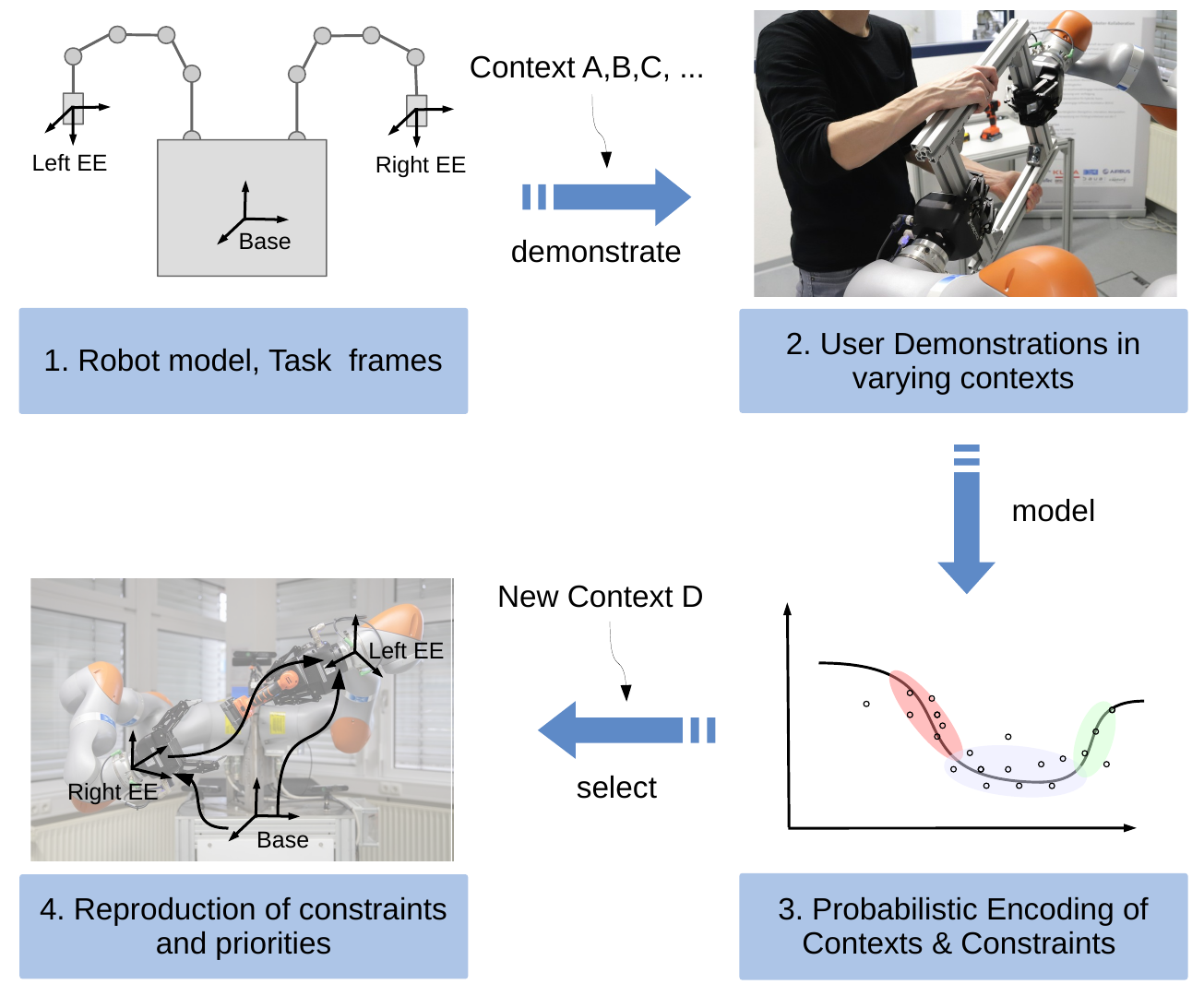}
\caption{Approach overview: Learning context-adaptive task constraints from user demonstrations}
\label{fig:wbc_learning}
\end{figure}

Here, we propose an approach that automatically derives task constraints from data recorded in user demonstrations. By recording the data in varying contexts we are able to generalize task constraints to novel situations. Figure~\ref{fig:wbc_learning} shows the general idea of the approach. 

\begin{enumerate}
\item We assume that the kinematic model of the robot is known, as well as  a number of task-relevant coordinate frames that have to be selected by the user in advance (e.g., the robot base, end effector or the coordinate frame of a certain object). We refer to these coordinate systems as \textit{task frames}, according to~\cite{Finkemeyer2004}. 
\item We perform $D$ user demonstrations in the form of kinesthetic teaching for each context. In general, the context may vary with respect to time. However, here we assume that the context remains constant throughout a single demonstration. In each demonstration we record the task constraints $(\mathbf{x}(t),\mathbf{v}(t))$ for each pair of task frames. Both, the pose $\mathbf{x}$ and twist $\mathbf{v}$ are represented as 6D arrays (see section~\ref{sec:learning_preprocessing} for details).

\item We model the joint probability distribution of task constraints and context variables as a Dirichlet Process Gaussian Mixture Model (DPGMM)
\item We reproduce task constraints and their respective priorities in a novel, previously unseen context using Gaussian Mixture Regression (GMR)
\end{enumerate}

In the following sections, we provide more detailed explanations of our approach. 

\subsection{Representation of Context}\label{sec:context_representation}

In our approach, we describe the context in the form of a \textit{context vector} $\mathbf{\kappa} \in \mathbb{R}^{C}$, where $C$ is the number of context variables. Currently, the user has to specify the context variables manually for each demonstration. The context variables can be real-valued, e.g., the size of an object, or categorical, e.g., whether an object is allowed to be tilted or not. In case of categorical variables we use one-hot encoding to model the different categories. Table~\ref{tab:context_example} shows an example of context variables, which describe the width of the manipulated object, whether the object may be tilted or not  and whether it should be manipulated with the right or left arm. The context vectors for these two cases would be $\kappa_1=(0.3,1,0,1)^T$ and $\kappa_2=(0.5,0,1,0)^T$.

\begin{table}[t]

\centering
    \footnotesize
	 \begin{tabular}{ccccc}
  \midrule
     Context \# & Object width & Allow Tilt & Right Arm & Left arm \\
   \midrule 
   1 & 0.3m & 1 & 0 & 1 \\
   2 & 0.5m & 0 & 1 & 0\\
  \midrule
 \end{tabular}
 \caption{Example context variables}
 \label{tab:context_example}
\end{table}

\subsection{Data Preprocessing}\label{sec:learning_preprocessing}
After recording, we first re-sample and temporally align all data streams. The time variable is normalized to $[0,1]$ to make the trajectories invariant with respect to time and linear scaling.

Since we use regression methods for reproduction of the demonstrated tasks we have to convert the rotational part of the pose trajectories to a suitable representation first. Euler angles are not unique, suffer from gimbal lock and have a discontinuous representation space, i.e. they wrap around $2\pi$. Thus, they are not well suited for regression. Orthogonal $3 \times 3$ rotation matrices have a continuous representation space, but are unfortunately over-parameterized. Also, during regression, the orthogonality constraint has to be enforced, e.g., by the means of a Gram-Schmidt orthonormalization process. Quaternions are not unique, discontinuous and the unit-length constraint has to be enforced during training. Thus, we decide to represent rotations as elements of the Lie algebra $so(3)$, which is the tangent space of $SO(3)$, the space of $3 \times 3$ orthogonal rotation matrices. An arbitrary element $\mathbf{R} \in SO(3)$ can be mapped to this 3-dimensional representation using the logarithmic map~\cite{lynch2017}:

\begin{equation}
\text{log}(\mathbf{R}) = [\mathbf{\hat{\omega}}] \theta 
\end{equation}

with

\begin{eqnarray}
\theta &=& \cos^{-1}(\frac{1}{2}(\text{tr}(\mathbf{R})-1))), \quad \theta \in [0,\pi]\\
\left[\mathbf{\hat{\omega}}\right]  & = & \frac{1}{2\sin(\theta)}(\mathbf{R}-\mathbf{R}^T), \quad \vert \text{tr}(\mathbf{R})\vert \neq 1
\end{eqnarray}

where $[\mathbf{\hat{\omega}}]$ is the skew-symmetric matrix form of the unit rotation axis $\mathbf{\hat{\omega}}$ and $\theta$ is the rotation angle for a given $\mathbf{R}$. The \textit{rotation vector} $\mathbf{\hat{\omega}}\theta$ gives us a 3D-representation of rotations. When restricting the rotation angle to $\theta \in [0,\pi]$, this representation will be unique (see e.g.,~\cite{Hartley2013}). However, when $\theta=0$ or $\theta=\pi$, the rotation axis inverts its sign. Thus, we have to handle these cases explicitly: First we ensure that the orientation trajectory starts in the upper half of $SO(3)$ ($\hat{\omega}_z \geq 0$). Then we walk through each data point in the recorded trajectory and apply $\mathbf{\hat{\omega}}^* = -\mathbf{\hat{\omega}}$ and $\theta^* = (2\pi-\theta)$ for the remaining elements whenever $\mathbf{\hat{\omega}}$ inverts its sign. As a result, we get a  continuous 3D-representation of our orientation data. 

The advantage of using $so(3)$ elements to represent rotations is that averaging of these elements is a linear operation just as it is for scalars and 3-dimensional position vectors, when the previously mentioned boundary cases are considered properly. Moreover, since we want to estimate the soft task priorities from the variability in the user demonstrations and the task weights in~(\ref{eq:opt_problem}) are six-dimensional (three entries correspond to the linear and angular velocity, respectively), we require a 3-dimensional representation of the orientation.

In summary, we represent the demonstrated motion in terms of time-varying 6-dimensional task constraints $(\mathbf{x}(t),\mathbf{v}(t))$, where $\mathbf{x} = (x,y,z,\phi,\theta,\psi)^T$ and $\mathbf{v} = (\dot{x},\dot{y},\dot{z},\dot{\phi},\dot{\theta},\dot{\psi})^T$. Each of these pairs describes the relative pose/twist between two task frames and can be used as input to the controller in Equation~(\ref{eq:pose_control}). The controller output represents a task constraint in Equation~(\ref{eq:opt_problem}) with 6 constraint variables, respectively. As a final preprocessing step, we normalize the complete data set to have zero mean and unit variance. 

After preprocessing we have, for each context, a normalized dataset $\mathbf{\xi} = [\mathbf{\mathcal{K}},\mathbf{X}, \mathbf{V}]$ with context data $\mathbf{\mathcal{K}}(t) \in \mathbb{R}^{D \cdot S \times C}$ and pose/twist trajectories $\mathbf{X}(t), \mathbf{V}(t) \in \mathbb{R}^{D \cdot S \times L}$. Here $t \in [0,1]$ is the normalized time variable, $D$ is the number of performed user demonstrations per context, $S$ the number of samples per experiment, $C$ the number of context variables and $L$ the number of constraint variables. The number of constraint variables  depends on the number of selected task frames $F$ as follows: $L = \frac{3F!}{(F-2)!}$ (e.g., for $F=3$, we have 18 constraint variables). Since $L$ strongly grows with $F$, the problem quickly becomes intractable for large $F$, so the task frames should be selected with care.

\begin{figure}
\centering
\includegraphics[width=\linewidth]{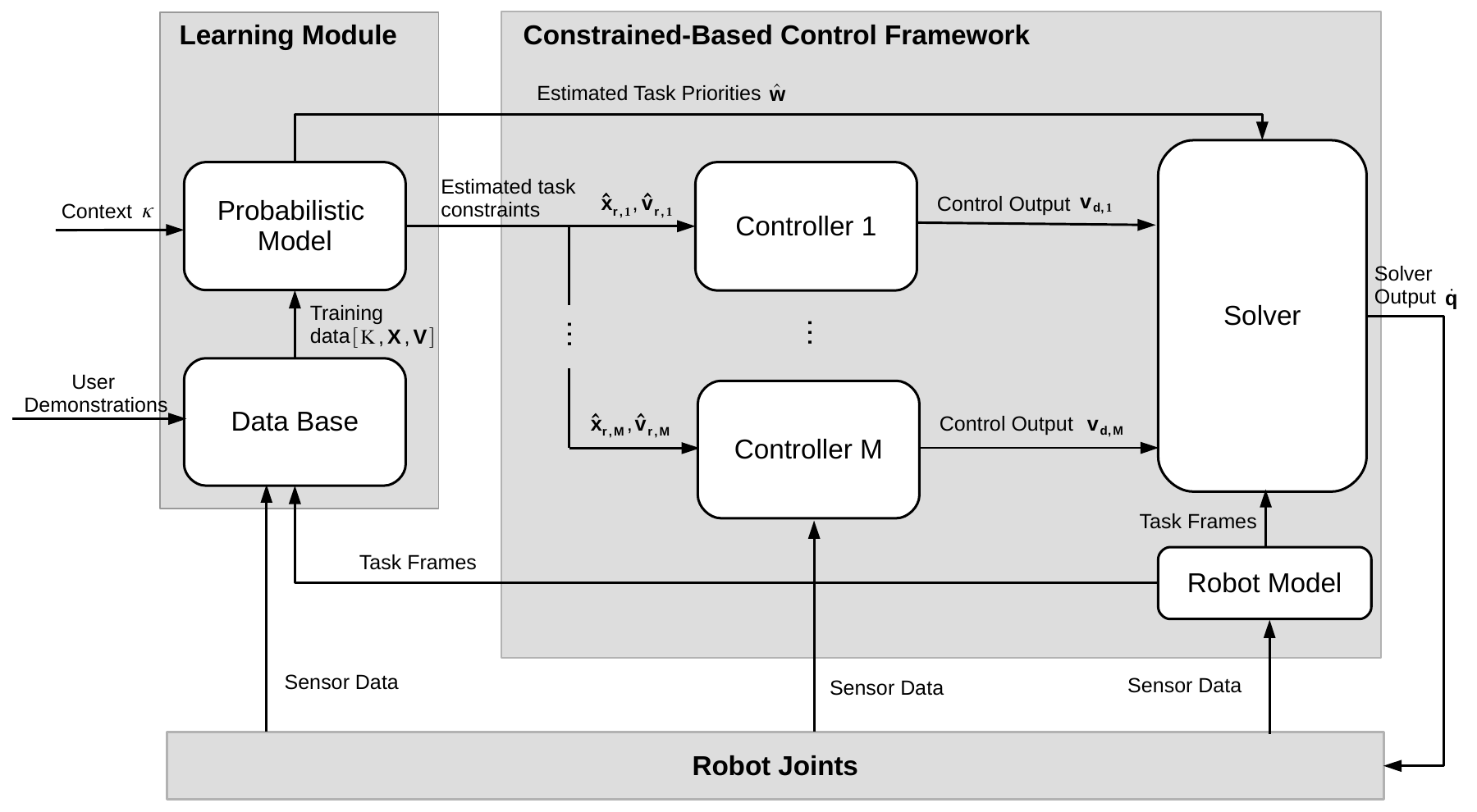}
\caption{Overview of the control framework including learning of context-adaptive task constraints. Each controller implements Equation~(\ref{eq:pose_control}), while the solver implements Equation~(\ref{eq:opt_problem})}
\label{fig:wbc_overview}
\end{figure}

\subsection{Estimation of Task Constraints and Priorities}\label{sec:constraint_estimation}

We want to estimate the task constraints $(\mathbf{x}(t),\mathbf{v}(t))$  and the respective soft task priorities $\mathbf{w}(t)$ that are required to reproduce the demonstrated task in a given context $\mathbf{\kappa}$. Figure~\ref{fig:wbc_overview} shows an overview of the framework including the learning module. During user demonstrations, we record the task constraints for each pair of task frames and store them in a data base. From a training data set $[\mathbf{\mathcal{K}}, \mathbf{X}, \mathbf{V}]$ we learn a probabilistic model. Given a certain context $\kappa$, the model estimates the task constraints $(\mathbf{x}(t), \mathbf{v}(t))$ and associated soft task priorities $\mathbf{w}(t)$ that are required to fulfill a certain task. Both, task constraints and priorities, are fed into the constraint-based control framework, which computes the joint velocities that comply with these constraints and sends them to the robot joints. 

For estimating task constraints, we learn the joint distribution of context and motion variables $\mathcal{P}(\mathbf{v},\mathbf{x},\mathbf{\kappa})$ as a Dirichlet Process Gaussian Mixture Model (DPGMM) from the recorded data. The model parameters $\{\mathbf{\pi}_k,\mathbf{\mu}_k, \mathbf{\Sigma}_k\}_{k=1}^K$ of the GMMs are trained using variational inference, where $K$ is the number of mixture components, $\mathbf{\pi}_k$ are the mixing weights, $\mathbf{\mu}_k$ the means and $\mathbf{\Sigma}_k$ the covariance matrices of the Gaussian distributions. In a DPGMM the mixing weights $\mathbf{\pi}_k$ are modeled as a Dirichlet Process, so that the effective number of mixture components can be inferred from data. In practice only an upper bound for the number of mixtures must be selected and the algorithm will set some of the mixture weights to near zero. 

Reproduction of the task constraints and the respective soft task priorities is then performed in an iterative manner: Starting from an initial pose\footnote{Note that $\mathbf{x}_0$ contains the relative poses of all pairs of task frames, stacked vertically. E.g., if the number of task frames is 3, the dimension of $\mathbf{x}_0$ is 18.} $\mathbf{x}_t=\mathbf{x}_0$, we estimate the twist $\hat{\mathbf{v}}_t$  from the conditional distribution $\mathcal{P}(\mathbf{v}|\mathbf{x}_t,\kappa)$ using Gaussian Mixture Regression (GMR)~\cite{Calinon2007} and integrate once to get the corresponding pose estimate:

\begin{equation}
\label{eq:pose_estimate}
\hat{\mathbf{x}}_{t+1} = \mathbf{x}_t + \hat{\mathbf{v}}_t \cdot \Delta t
\end{equation}

where $\Delta t$ is the sample time in seconds. Given the estimated twist, we can compute the conditional distribution $\mathcal{P}(\mathbf{x}|\mathbf{\hat{v}}_{t},\mathbf{\kappa})$. From this distribution with parameters $\{\mathbf{\pi}_k,\mathbf{\mu}_k, \mathbf{\Sigma}_k\}_{k=1}^K$ we can get an estimate of the  variance for each constraint variable by collapsing the multi-variate Gaussian distribution to a single Gaussian as follows:

\begin{equation}
\label{eq:single_gaussian}
\mathbf{\mu} = \sum_{i=1}^K \pi_i\mathbf{\mu}_i, \quad \mathbf{\Sigma} = \sum_{i=1}^K \pi_i(\Sigma_i + \mu_i\mu_i^T + \mathbf{\mu}\mathbf{\mu}^T)
\end{equation}

Note that we omit the time index for the sake of readability here. From the covariance matrix $\mathbf{\Sigma}$, we compute the task weights $\mathbf{w}_t$ as follows:

\begin{equation}
\label{eq:weight_estimate}
w_j = 1 - \left(\frac{\mathbf{\sigma}_j^2}{\bar \sigma^2}\right), \quad \forall j
\end{equation} 
where $\sigma_j^2$ are the diagonal entries of $\mathbf{\Sigma}$ and $\bar \sigma^2$ is the maximum variance over all constraint variables. Finally, we set  $\mathbf{x}_t = \hat{\mathbf{x}}_{t+1}$, estimate the next twist and so on. This process is repeated until converging to the target pose.

\begin{algorithm}[t]
\footnotesize
\caption{Reproduction of Task Constraints and Soft Task Priorities}
\begin{algorithmic}[1]
\State Given: Joint distribution $\mathcal{P}(\mathbf{v},\mathbf{x},\mathbf{\kappa})$, Context $\mathbf{\kappa}$
\State Start at $t = 0$, start pose $\mathbf{x}_t=\mathbf{x}_0$
\While{$\|\mathbf{x}_e-\mathbf{x}_t\|_2 > \delta$ }
\State \textbf{1. Estimate twist}: 
\State \hspace{0.5cm}From $\mathcal{P}(\mathbf{v}|\mathbf{x}_t,\kappa)$ estimate $\hat{\mathbf{v}}_{t}$ using GMR
\State \textbf{2. Estimate pose} 
\State \hspace{0.5cm}Integrate once to get the corresponding pose estimate as in Equation~(\ref{eq:pose_estimate})
\State{\textbf{3. Estimate task weights}: 
\State \hspace{0.5cm}Compute $\mathcal{P}(\mathbf{x}|\mathbf{\hat{v}}_{t},\mathbf{\kappa})$ using the estimated twist $\mathbf{\hat{v}}_{t}$
\State \hspace{0.5cm}Compute task weights using  Equations~(\ref{eq:single_gaussian}) and (\ref{eq:weight_estimate}})
\State \textbf{4. Update}
\State \hspace{0.5cm}Set $\mathbf{x}_t = \hat{\mathbf{x}}_{t+1}$
\EndWhile 
\end{algorithmic}
\label{alg:reproduction}
\end{algorithm}

The procedure for estimating task constraints from GMM is summarized in Algorithm~\ref{alg:reproduction}. The input to the algorithm is the context vector $\mathbf{\kappa} \in \mathbb{R}^C$ and the initial relative poses for each pair of task frames $\mathbf{x}_{0}$. The output are the task constraints $(\mathbf{x}(t),\mathbf{v}(t))$, as well as the associated soft task priorities $\mathbf{w}(t)$. By using twist commands as variables, the acquired trajectories can be adapted with respect to varying starting points.

The advantage of GMR over other regression techniques is that it is able to generate smooth and continuous motions and that it provides information about the variance of the input data, which we require to estimate the soft task priorities. Furthermore, the time for regression is independent of the size of the data set, as GMR models the joint probability of the data, and then derives the regression function from the joint density model~\cite{Stulp2015}. Since we have quite large data sets we prefer GMM-GMR over other approaches that model the regression function directly like e.g., Gaussian Process Regression (GPR)~\cite{Rasmussen2004}. 

Since we perform $D$ different user demonstrations per context, each point in the trajectories can be assigned a variance $\sigma_\kappa(t)$, which describes the variability of the demonstrations in context $\mathbf{\kappa}$. In the algorithm, we use this variance to provide an estimate of the task priorities. The key idea is that a high variability in the user demonstrations corresponds to a low priority of the task constraints and vice versa. Figuratively, this means that a demonstrated motion with low variability throughout all demonstrations is "constrained" and thus very important for the performed task, while a high variability reflects less important parts of the task. When, for example, performing a task like polishing a table, the motion perpendicular to the table surface is constrained and a low variability will be perceived in that direction. Thus the corresponding task constraint is assigned a high priority. The motion parallel to the surface on the other hand is quite arbitrary and can be assigned lower priority, i.e. the motion must not be tracked very accurately.

\begin{figure}
\centering
\begin{subfigure}{0.49\linewidth}
\includegraphics[width=\columnwidth]{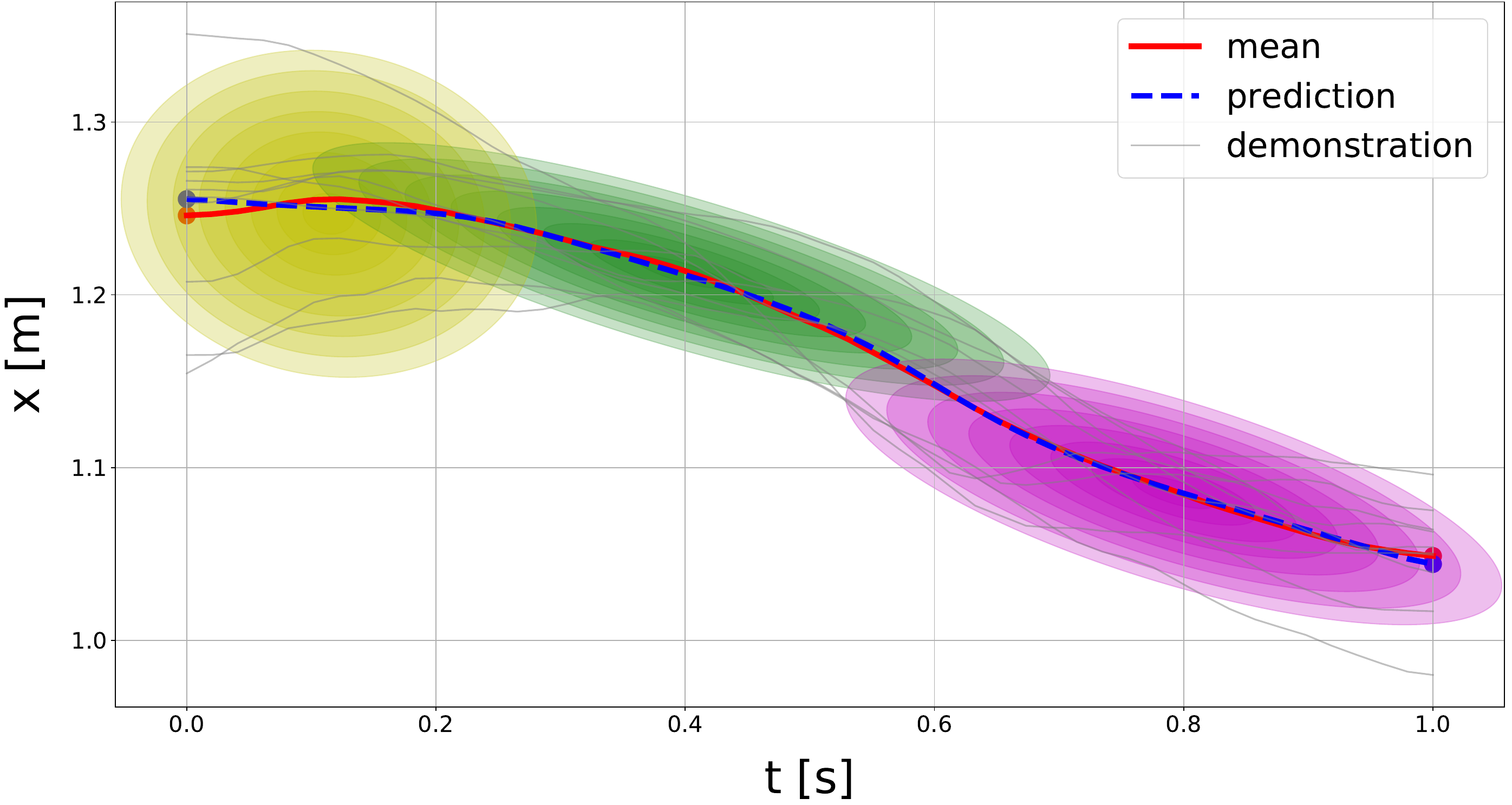}
\caption{Mixture Components ($K=3$)}
\label{fig:task_constraint_estimation_example_1}
\end{subfigure}
\begin{subfigure}{0.49\linewidth}
\includegraphics[width=\columnwidth]{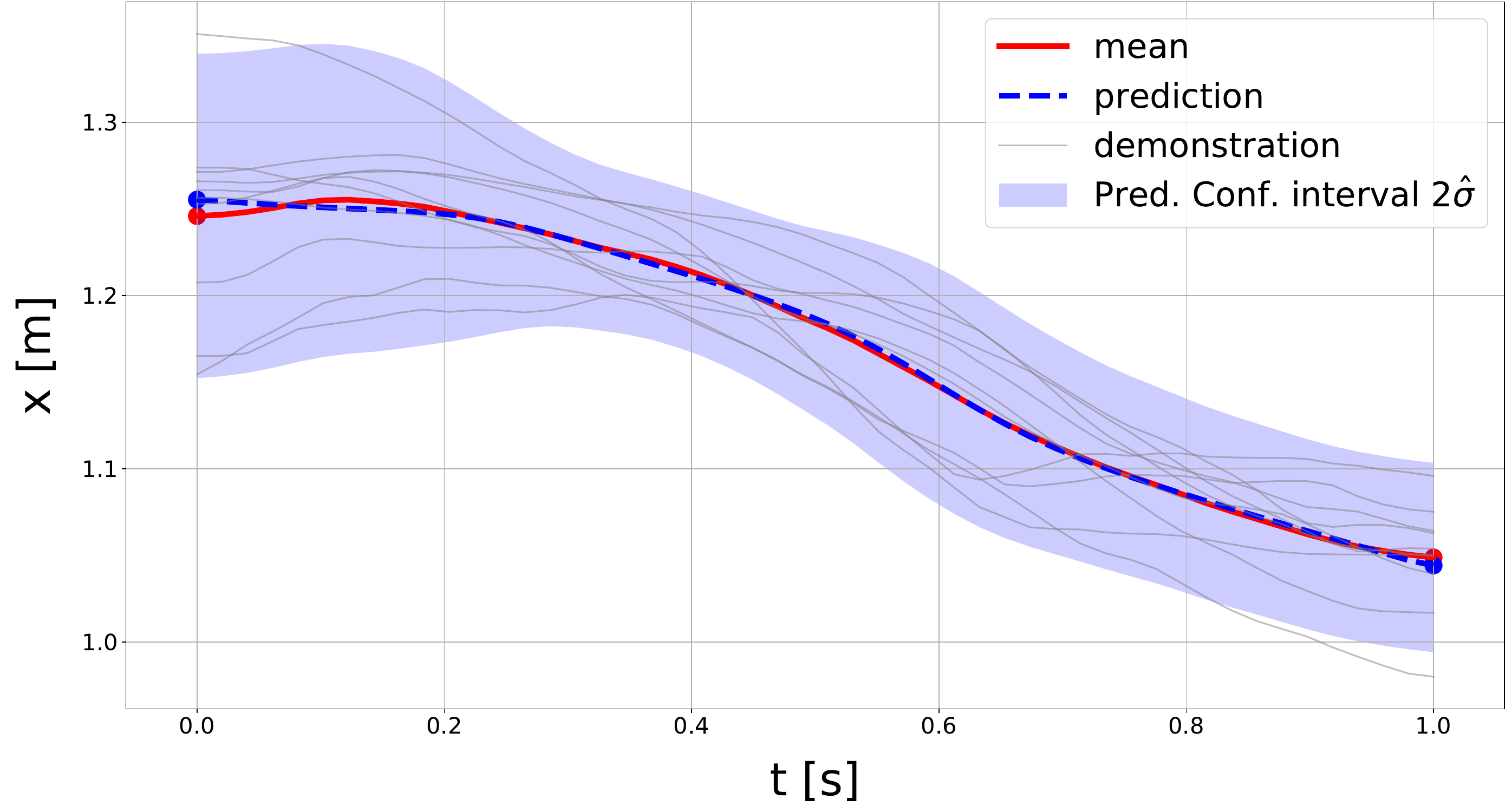}
\caption{Resulting Confidence Interval ($2\sigma$)}
\label{fig:task_constraint_estimation_example_2}
\end{subfigure}
\caption{Example: Estimated task constraints (only x-position) and confidence interval, which is used for predicting the task weights.}
\label{fig:task_constraint_estimation_example}
\end{figure}

As an example, Figure~\ref{fig:task_constraint_estimation_example} illustrates the reproduction of a motion (only x-position) in a fixed context. Figure~\ref{fig:task_constraint_estimation_example_1} shows the mean and spread of $K=3$ mixture components fitted to $D=10$ different user demonstrations, the predicted trajectory using GMR and the mean trajectory from the user demonstrations. Figure~\ref{fig:task_constraint_estimation_example_2} shows the resulting confidence interval $2\sigma$, which is used to estimate the task weights according to Equation~(\ref{eq:weight_estimate}).

\subsection{Generalization to unknown Contexts}\label{sec:learning_generalize}

In order to achieve generalization capabilities with respect to previously unseen situations we perform the demonstrations under multiple variations of the given task. We refer to these variations as \textit{context changes} here. As described before, the context is described by a real-valued vector $\mathbf{\kappa}$, where categorical variables are modeled using one-hot encoding. Previously introduced approaches like the one described in~\cite{Calinon2016} focus on generalization over different start or target positions for a given task. Here, we want to additionally deal with more severe context changes, e.g., the size of the handled objects, whether to use a single arm or two arms for the given task or whether or not an object may be tilted during task execution. Such changes can be represented in our control approach by modifying task weights of particular constraints in an appropriate way. For example, if an object may be tilted during execution, the task weights corresponding to the rotational motion can be low, so that the remaining degrees of freedom can be used by the robot to perform additional tasks, like collision avoidance. 

One problem with Gaussian Mixture Models is the selection of the number of components or mixtures. If it is chosen too large, the resulting model may represent the training data accurately, but does not generalize well to previously unseen samples. DPGMM allows to infer the number of active components from data, with the downside of requiring additional hyper-parameters to tune. The most important hyper-parameter is the weight concentration prior $\gamma$. A small value of $\gamma$ sets most component's weights to zero, which leads to a small number of active components in practice. A large value of $\gamma$ produces an equally distributed weight concentration over all components, which corresponds to having a large number of active components. 
To achieve best generalization capabilities, we optimize the weight concentration prior and other hyper-parameters of the DPGMM using grid search. The training data is selected using leave-one-out cross validation, where we use the data from each context as a hold out set once in each split and train on the $C-1$ remaining contexts. Finally, we test the resulting model in a context that the model has not seen before. 

\begin{figure*}[t!]
\centering
\begin{subfigure}[t]{0.3\textwidth}
\includegraphics[width=\textwidth]{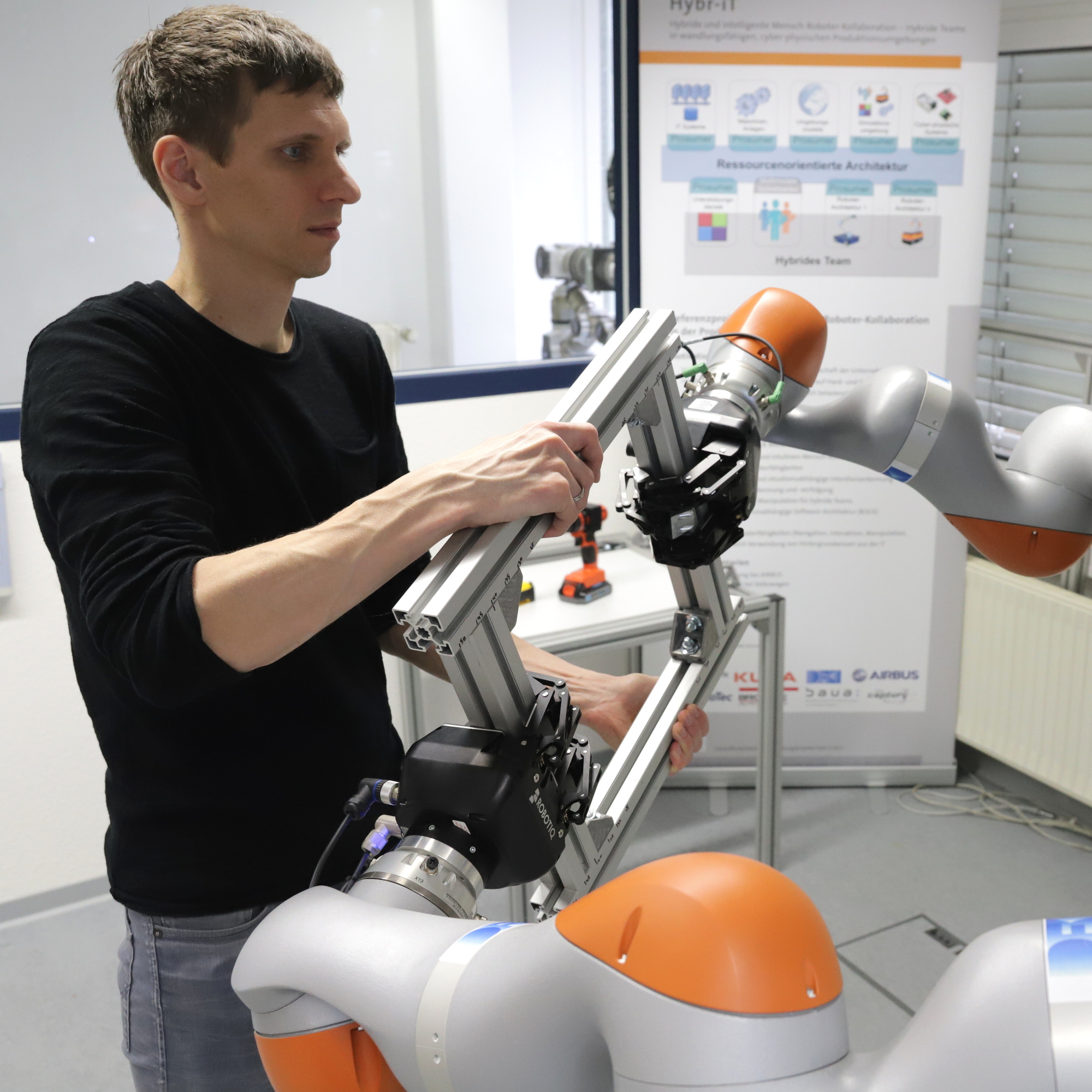}
\caption{\textit{Rotate object}: Rotating an object by 90° degrees}
\label{fig:demonstrations_rotate}
\end{subfigure}
\hspace{0.01\textwidth}
\begin{subfigure}[t]{0.3\textwidth}
\includegraphics[width=\textwidth]{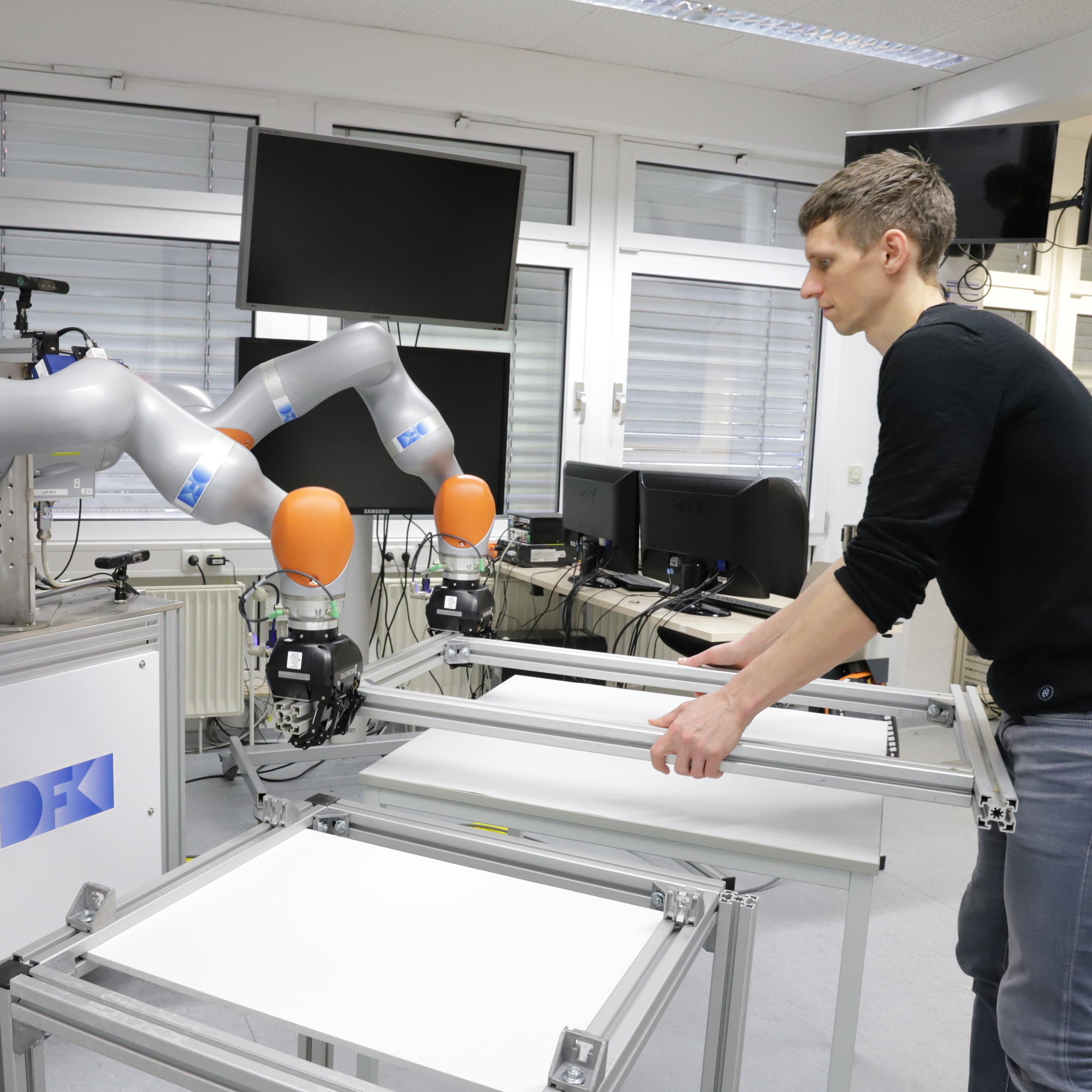}
\caption{\textit{Collaboration}: Collaborative transport of a bulky object}
\label{fig:demonstrations_carry}
\end{subfigure}
\hspace{0.01\textwidth}
\begin{subfigure}[t]{0.3\textwidth}
\includegraphics[width=\textwidth]{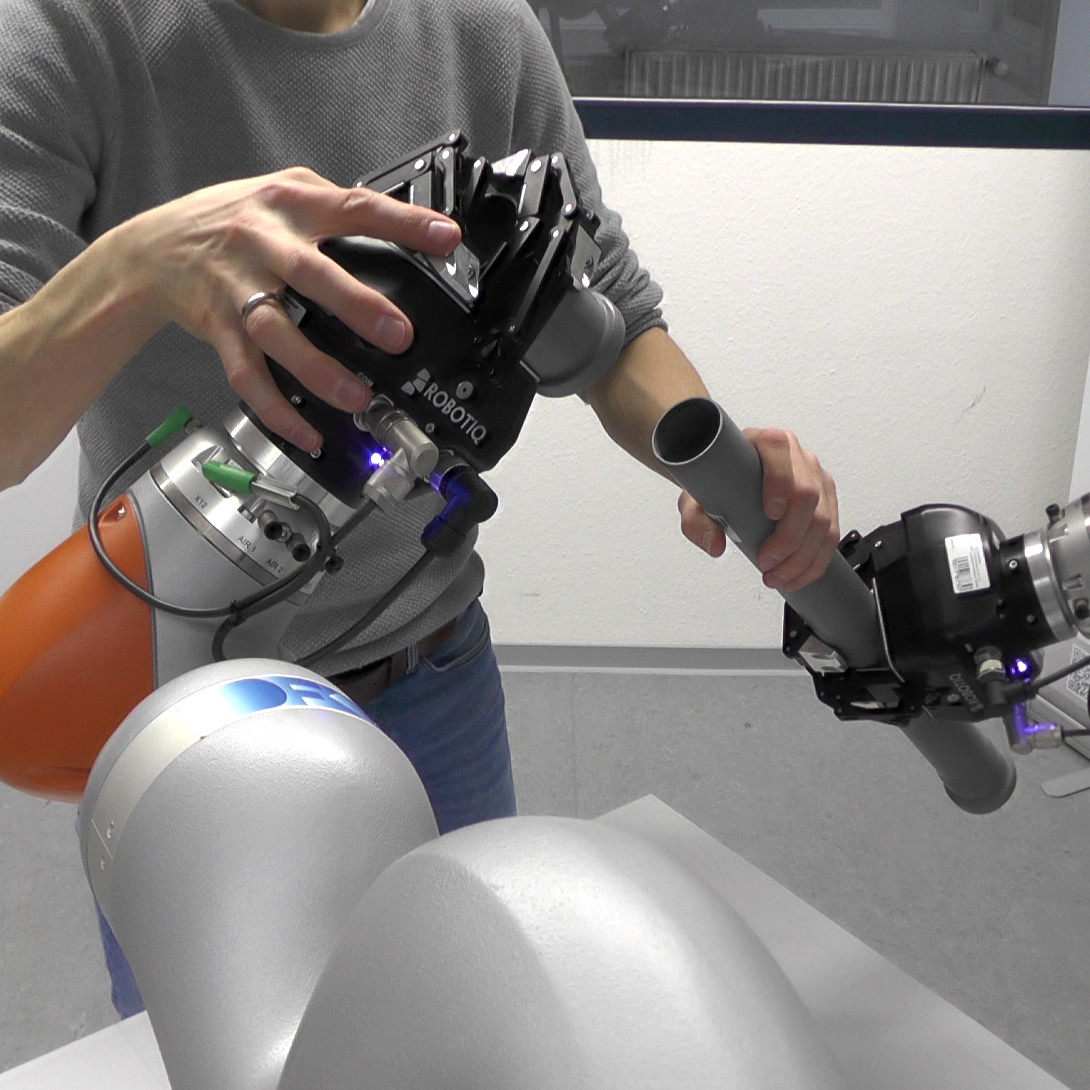}
\caption{\textit{Assembly}: Connecting a tube and a connector piece}
\label{fig:demonstrations_assemble}
\end{subfigure}
\caption{Kinesthetic teaching of dual-arm manipulation tasks}
\label{fig:demonstrations}
\end{figure*}

\begin{table*}
    \centering
    \begin{minipage}{\textwidth}
    \centering
    \footnotesize
\centering
\begin{subfigure}[t]{\columnwidth}
 \centering
 \begin{tabular}{clccccc}
  \midrule
     \#  & Name & OS & LA & RA & C\\
   \midrule 
   $R_{11}$ & Rot. $0.30m$ clockw. & 0.3 & 1 & 1 & 1\\
   $R_{12}$ & Obj. $0.35m$ clockw. & 0.35 & 1 & 1 & 1 \\
   $R_{13}$ & Obj. $0.40m$ clockw. & 0.4 & 1 & 1 & 1\\
   $R_{14}$ & Obj. $0.45m$ clockw. & 0.45 & 1 & 1 & 1 \\
   $R_{15}$ & Obj. $0.50m$ clockw. & 0.5 & 1 & 1 & 1\\
   \midrule 
   $R_{21}$ & Obj. $0.30m$ anticlockw. & 0.3 & 1 & 1 & 0   \\
   $R_{22}$ & Obj. $0.35m$ anticlockw. & 0.35 & 1 & 1 & 0   \\ 
   $R_{23}$ & Obj. $0.40m$ anticlockw. & 0.4 & 1 & 1 & 0   \\
   $R_{24}$ & Obj. $0.45m$ anticlockw. & 0.45 & 1 & 1 & 0   \\ 
   $R_{25}$ & Obj. $0.50m$ anticlockw. & 0.5 & 1 & 1 & 0   \\
   \midrule 
   $R_{31}$ & Obj. $0.50m$ left arm anticlockw. & 0.5 & 1 & 0 & 0   \\
   $R_{32}$ & Obj. $0.50m$ left arm clockw.  & 0.5 & 1 & 0 & 1  \\
   \midrule 
   $R_{41}$ & Obj. $0.50m$ right arm clockw. & 0.5 & 0 & 1 & 1  \\
   $R_{42}$ & Obj. $0.50m$ right arm anticlockw.  & 0.5 & 0 & 1 & 0  \\
  \midrule
 \end{tabular} 
  \caption{\emph{Rotate Object}}
  \label{tab:contexts_rotate_object}
 \end{subfigure}
    \end{minipage}%
    
    \begin{minipage}{\textwidth}
    \centering
\begin{subfigure}[t]{\columnwidth}
 \centering
    \footnotesize
	 \begin{tabular}{clccc}
  \midrule
     \#  & Name & AT & LA & RA \\
   \midrule 
   $C_{11}$ & Collab. no tilt & 0 & 1 & 1 \\
   $C_{12}$ & Collab. with tilt & 1 & 1 & 1 \\
   \midrule
   $C_{21}$ & Collab. no tilt left arm & 0 & 1 & 0 \\
   $C_{22}$ & Collab. with tilt left arm & 1 & 1 & 0 \\
   \midrule
   $C_{31}$ & Collab. no tilt right arm & 0 & 1 & 0 \\
   $C_{32}$ & Collab. with tilt right arm & 1 & 1 & 0 \\
  \midrule
 \end{tabular}
  \caption{\emph{Collaboration}}
 \end{subfigure}  \\ \vspace{0.3cm}

\begin{subfigure}[t]{\columnwidth}
 \centering
 \begin{tabular}{clcc}

  \midrule
     \#  & Name & LA & RA \\
   \midrule 
   $A_{11}$ & Assembly & 1 & 1 \\
   $A_{21}$ & Assembly left arm & 1 & 0 \\
   $A_{31}$ & Assembly right arm & 0 & 1 \\
  \midrule
 \end{tabular}
  \caption{\emph{Assembly}}
 \end{subfigure}
    \end{minipage}%
    \caption{Contexts and context variables used for experimental evaluation,  OS - Object Size, C - Clockwise rotation, LA/RA - Left Arm/Right Arm, AT - Allow Tilt}
     \label{tab:contexts}
\end{table*}

\section{EXPERIMENTAL RESULTS} \label{sec:results}

We evaluate our approach by the means of 3 different manipulation tasks:\\
\emph{Rotate Object} The robot rotates a rigid object by $90$ degrees (Figure~\ref{fig:demonstrations_rotate}). We vary the start pose, the width of the object (between $0.3m$ and $0.5m$), the rotation direction (clockwise/anticlockwise) and whether both robot arms or a single arm (left arm/right arm) is used for execution. In total we get 14 different contexts, parameterized by $C=4$  context variables. The user demonstrations of this task are illustrated in the accompanying video \url{anc/01_pbd_rotate_panel.mp4}{01\_pbd\_rotate\_panel.mp4}.\\
\emph{Collaboration} The robot carries a bulky object in collaboration with a human (Figure~\ref{fig:demonstrations_carry}). We vary the start pose, whether or not the object may be tilted during transport and whether both robot arms or a single arm (left arm/right arm) is used for the experiment. We obtain data in 6 different contexts, parameterized by $C=3$ context variables.  The user demonstrations of this task are illustrated in the accompanying video \url{anc/02_pbd_collaboration.mp4}{02\_pbd\_collaboration.mp4}.\\
\emph{Assembly} The robot assembles a tube and a connector piece. We vary the start pose and whether both robot arms or a single arm (left arm/right arm) is used for the experiment.  Thus, we perform the task in 3 different contexts, parameterized by $C=2$ context variables. The user demonstrations of this task are illustrated in the accompanying video \href{anc/03_pbd_assembly.mp4}{03\_pbd\_assembly.mp4}.

A summary of all recorded contexts and the context variables can be found in Table~\ref{tab:contexts}.

The experiments are conducted on a stationary dual-arm robot consisting of two KUKA iiwa lightweight arms\footnote{https://www.kuka.com/en-us/products/robotics-systems/industrial-robots/lbr-iiwa}, each equipped with an Robotiq 3-finger gripper\footnote{https://robotiq.com/products/3-finger-adaptive-robot-gripper}. We select the base frame of the robot (denoted as \emph{Base}), as well as the end-effector frames of the two arms (denoted as \emph{Left EE} and \emph{Right EE}) as task frames. The resulting task constraints will be denoted as \emph{Base}-\emph{Left EE}, \emph{Base}-\emph{Right EE} and \emph{Left EE}-\emph{Right EE} in the following. Since we have three 6-dimensional Cartesian constraints, we get $L=18$ pose and $L=18$ twist variables, respectively. For each context, we perform $D=10$ experiments (with varying start pose). The recorded trajectories are re-sampled to contain $S=200$ samples each. 

\begin{figure*}[t!]
\centering
\begin{subfigure}{0.45\textwidth}
\includegraphics[width=\textwidth]{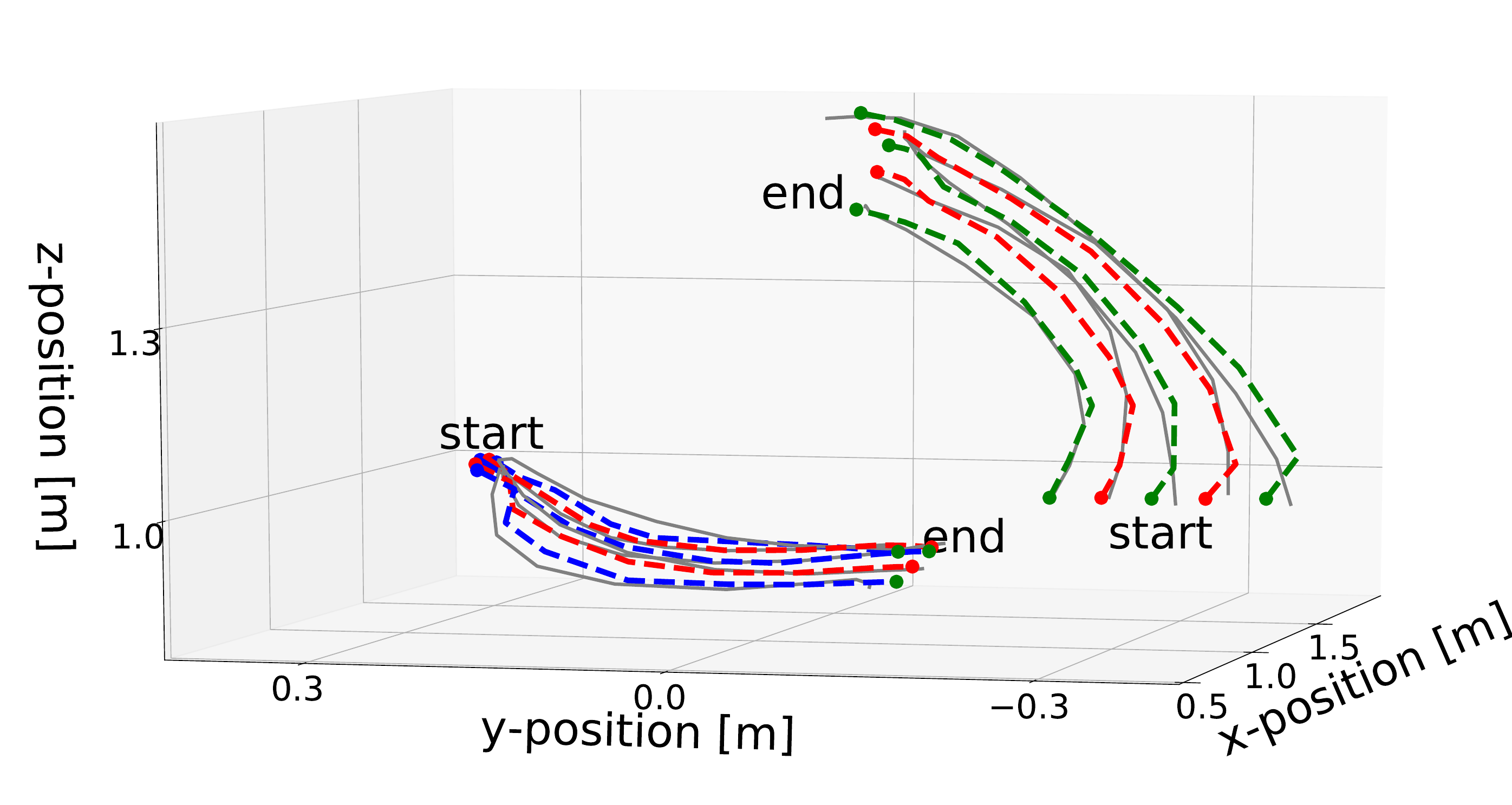}
\caption{Rotate Object: Reproduction with varying object size}
\label{fig:results_rotate_panel_varying_panel_size}
\end{subfigure}
\hspace{0.05\textwidth}
\begin{subfigure}{0.45\textwidth}
\includegraphics[width=\textwidth]{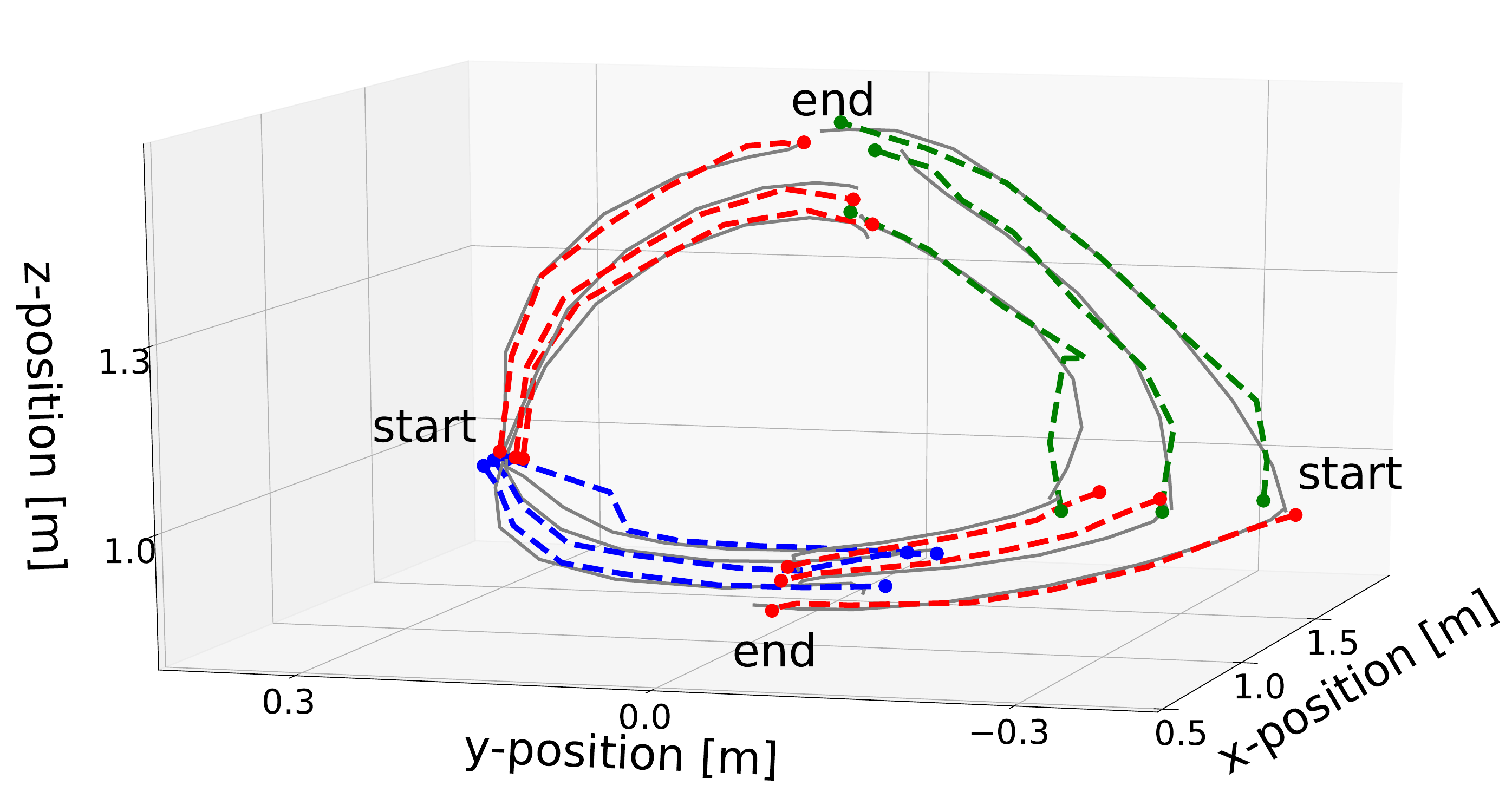}
\caption{Rotate Object: Reproduction with varying rotation direction}
\label{fig:results_rotate_panel_varying_rot_dir}
\end{subfigure}\\
\begin{subfigure}{0.45\textwidth}
\includegraphics[width=\textwidth]{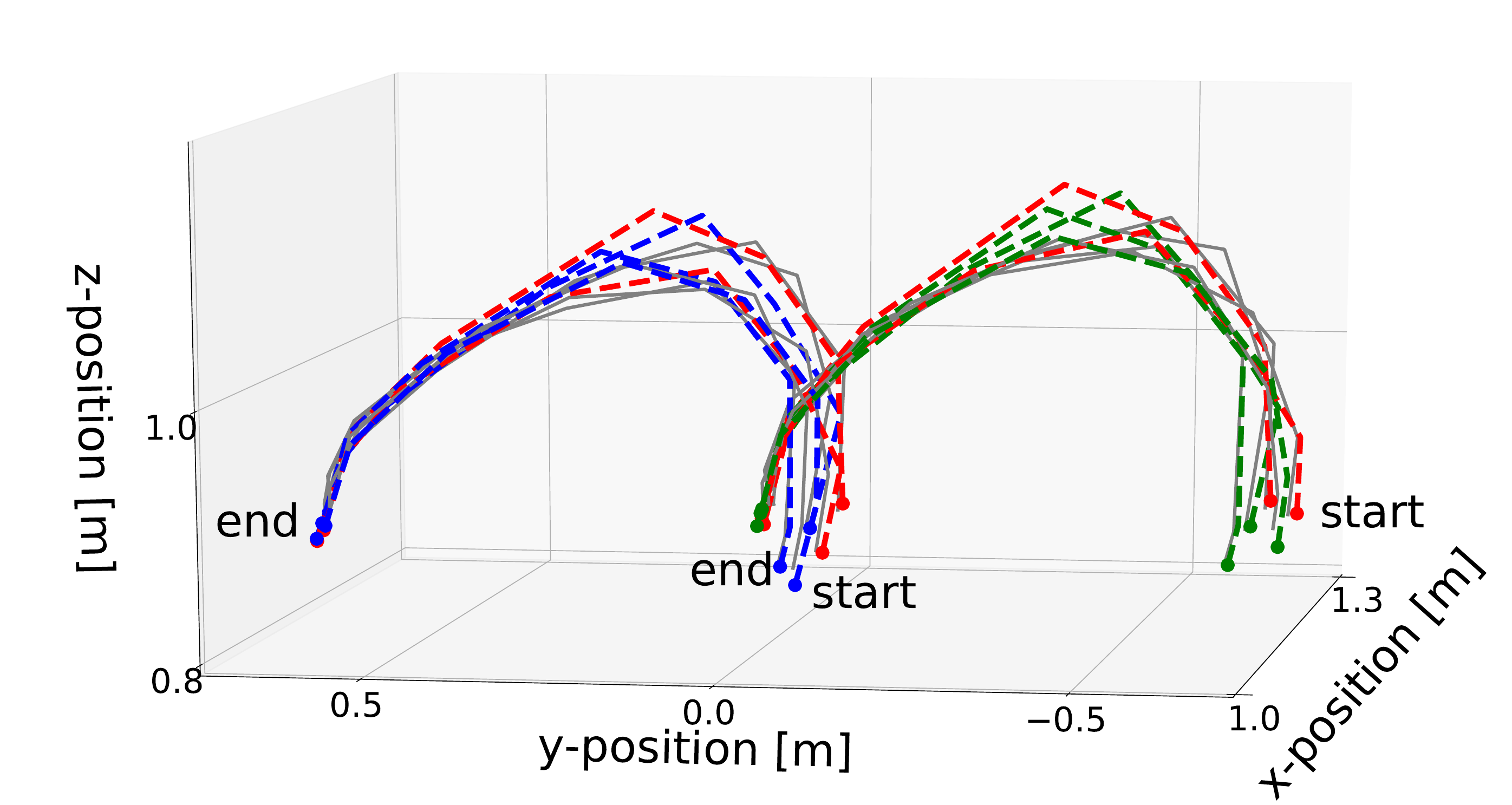}
\caption{Collaboration: Reproduction with varying start position}
\label{fig:results_carry_load_varying_start_pos}
\end{subfigure}
\hspace{0.05\textwidth}
\begin{subfigure}{0.45\textwidth}
\includegraphics[width=\textwidth]{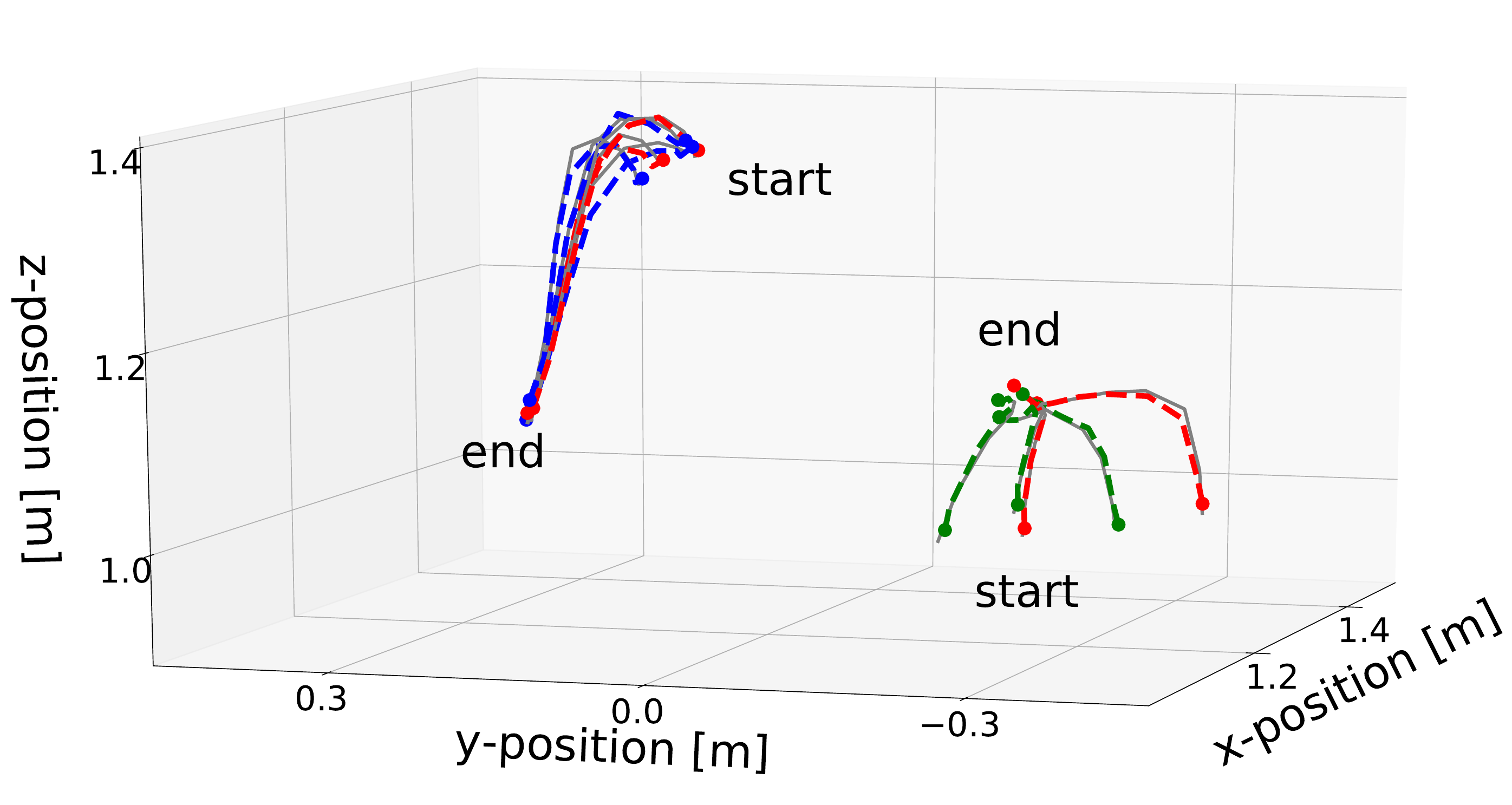}
\caption{Assembly: Reproduction with varying start position}
\label{fig:results_assembly_varying_start_pos}
\end{subfigure}
\caption{Results when reproducing task constraints in previously unseen context: \color{darkgray}Gray\color{black}: Mean of demonstrations, \color{blue}Blue Dashed\color{black}: Left Arm (constraint \textit{Base}-\textit{Left EE}), \color{ForestGreen}Green Dashed\color{black}: Right Arm (constraint \textit{Base}-\textit{Right EE}), \color{red}Red Dashed\color{black}: Reproduction in previously unseen context}
\label{fig:results_motion_constraints}
\end{figure*}

\begin{figure}
\centering
\includegraphics[width=0.4\linewidth]{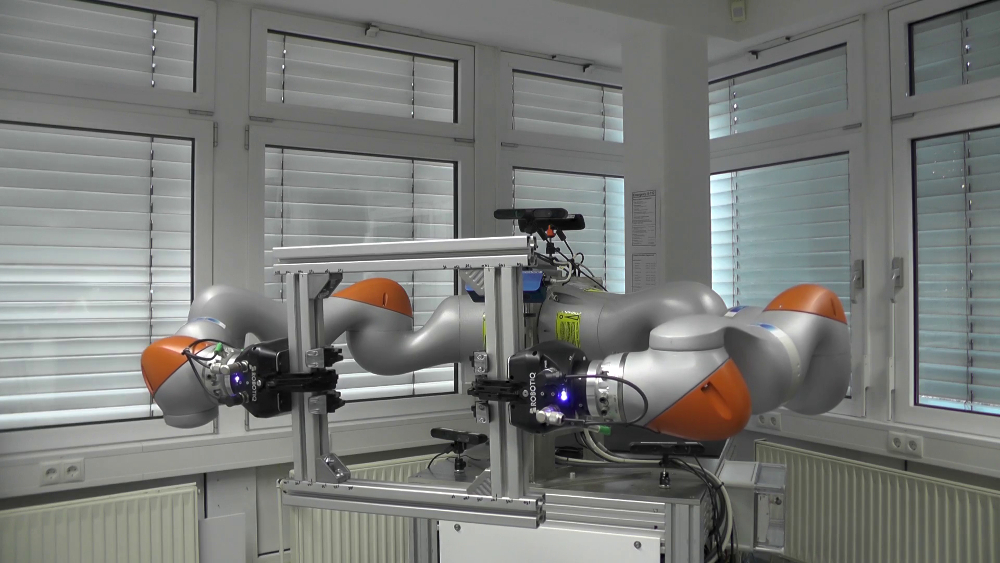}
\includegraphics[width=0.4\linewidth]{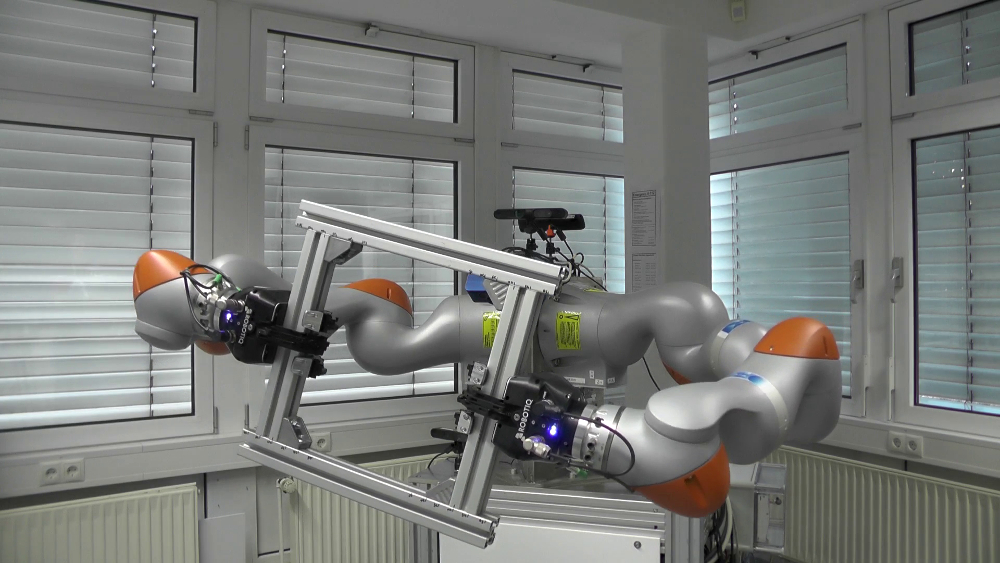}\vspace{0.1cm}\\
\includegraphics[width=0.4\linewidth]{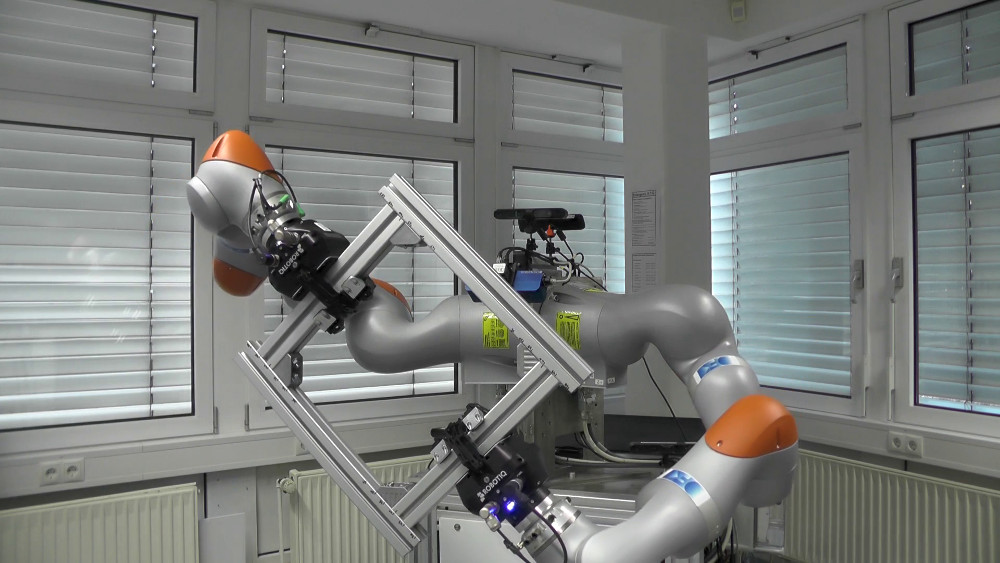}
\includegraphics[width=0.4\linewidth]{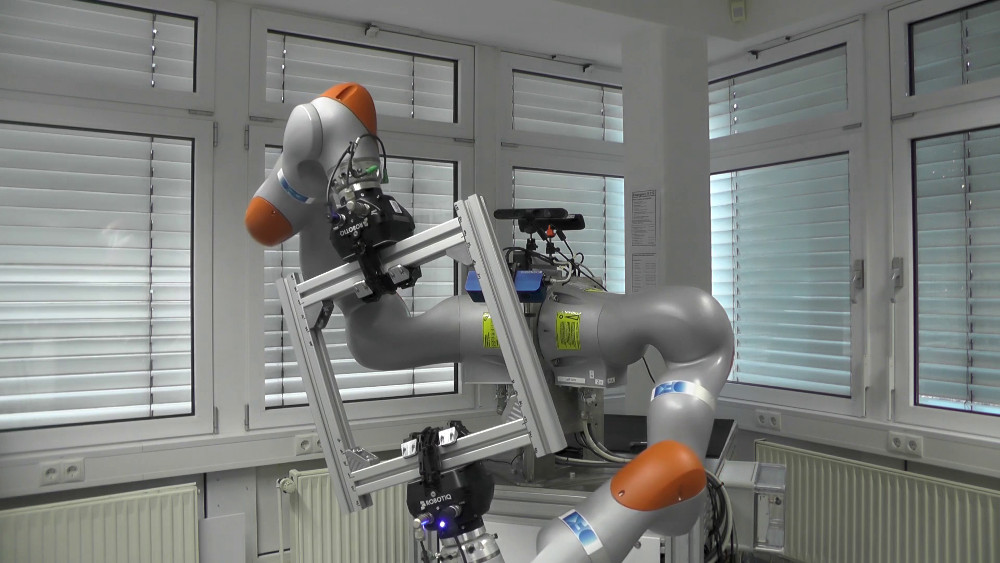}\vspace{0.1cm}\\
\includegraphics[width=0.4\linewidth]{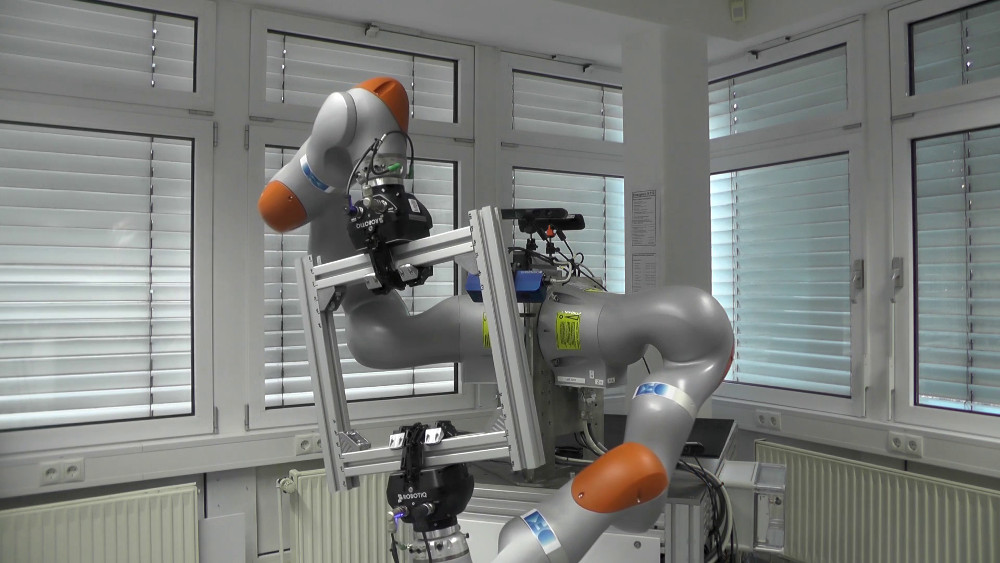}
\includegraphics[width=0.4\linewidth]{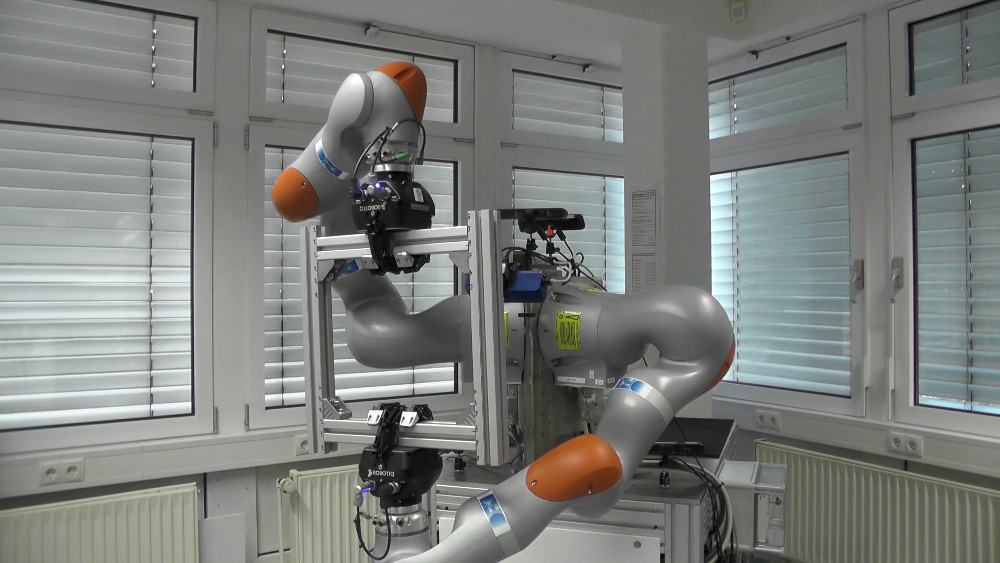}
\caption{Reproduction of the \textit{Rotate Object} task in context $R_{12}$}
\label{fig:results_rotate_panel}
\end{figure}

\begin{figure}
\centering
\includegraphics[width=0.4\linewidth]{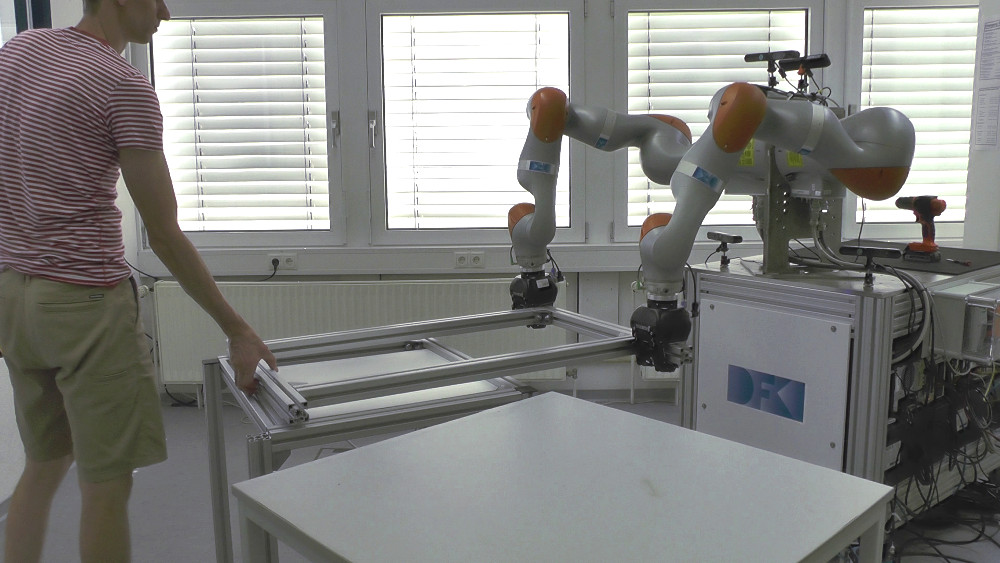}
\includegraphics[width=0.4\linewidth]{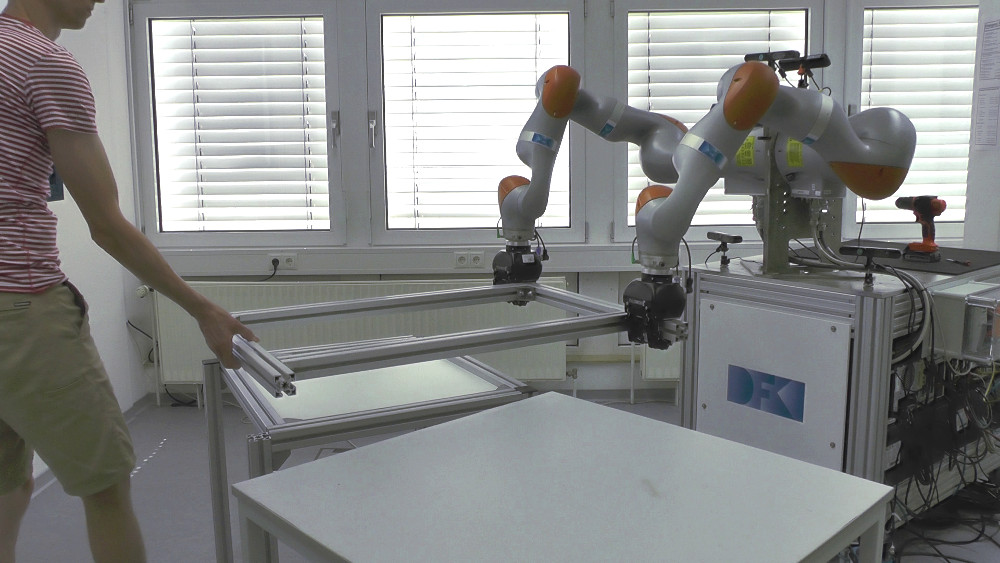}\vspace{0.1cm}\\
\includegraphics[width=0.4\linewidth]{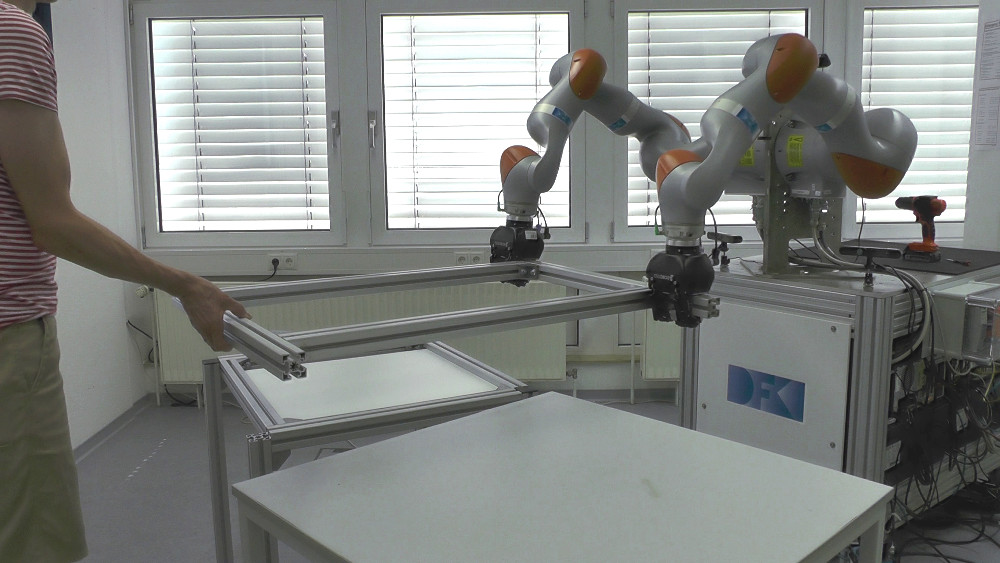}
\includegraphics[width=0.4\linewidth]{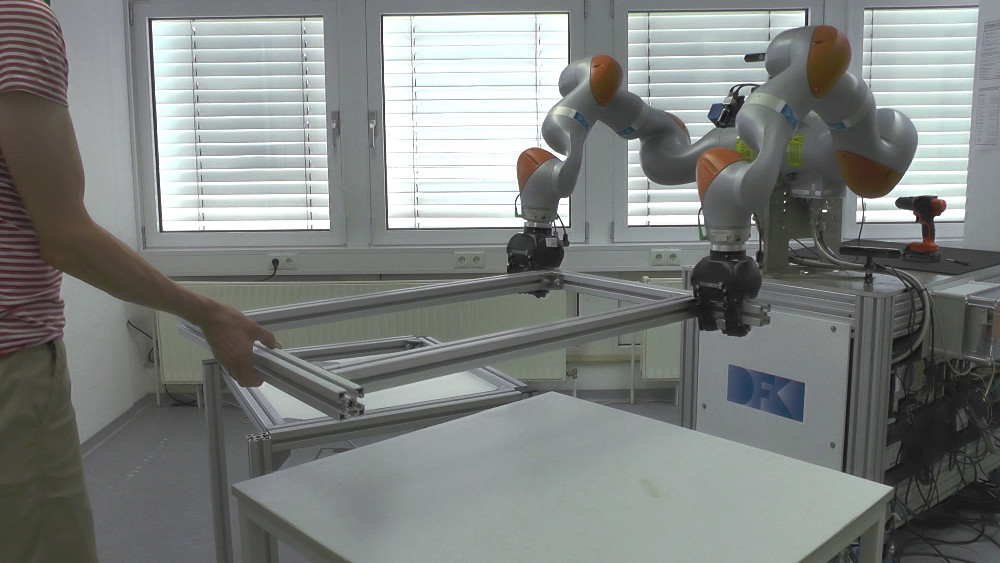}\vspace{0.1cm}\\
\includegraphics[width=0.4\linewidth]{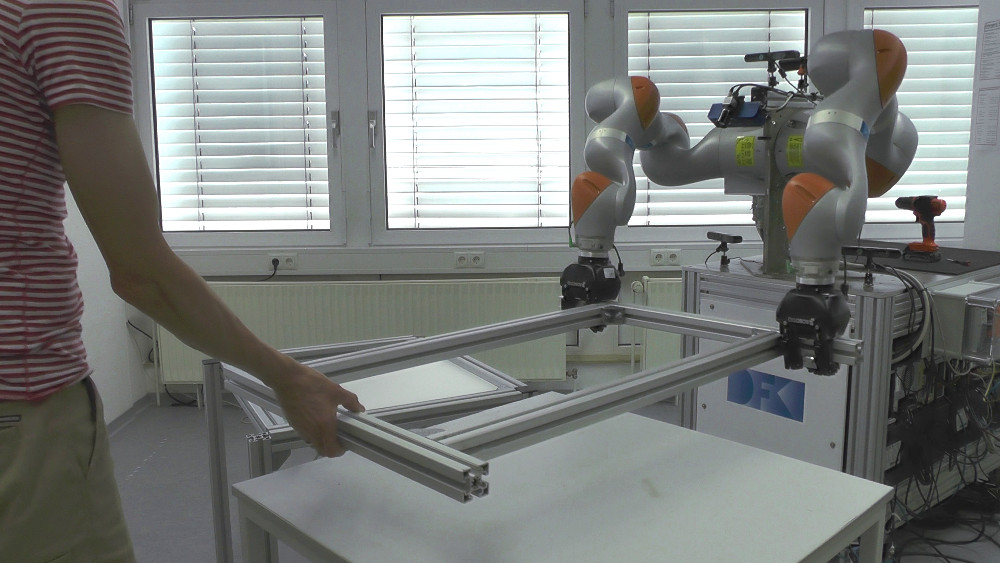}
\includegraphics[width=0.4\linewidth]{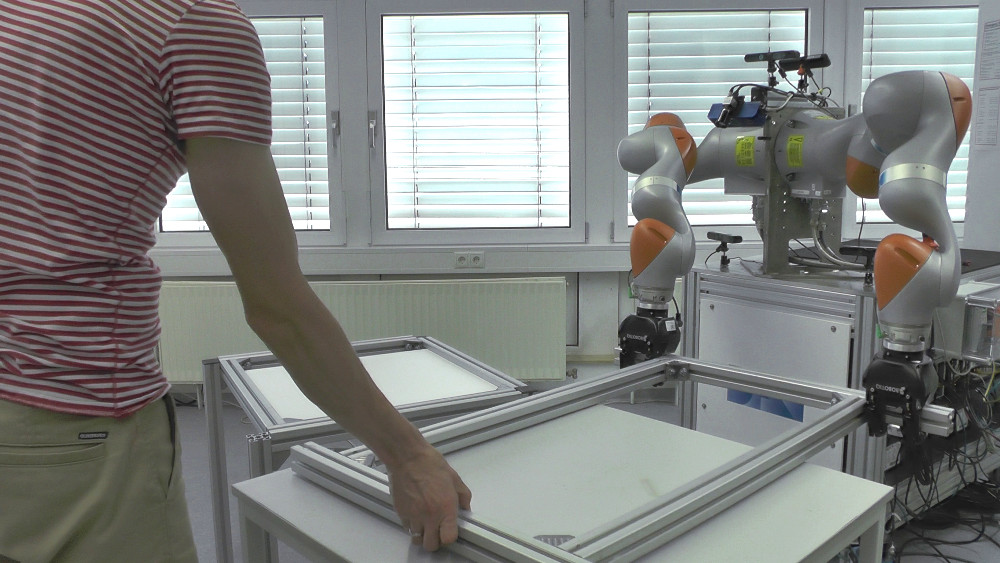}
\caption{Reproduction of the \textit{Collaboration} task in context $C_{11}$}
\label{fig:results_carry_load}
\end{figure} 

\begin{figure}
\centering
\includegraphics[width=0.4\linewidth]{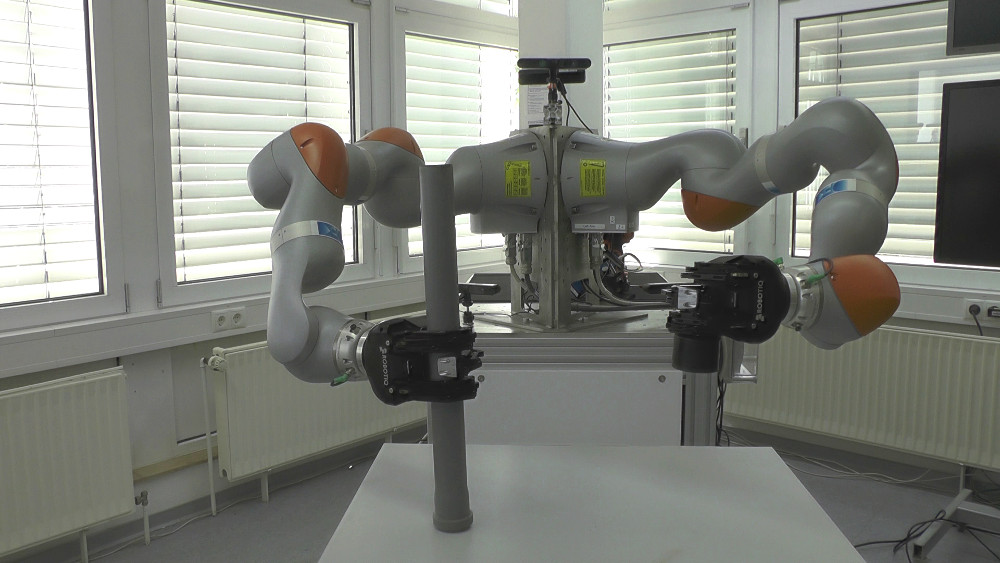}
\includegraphics[width=0.4\linewidth]{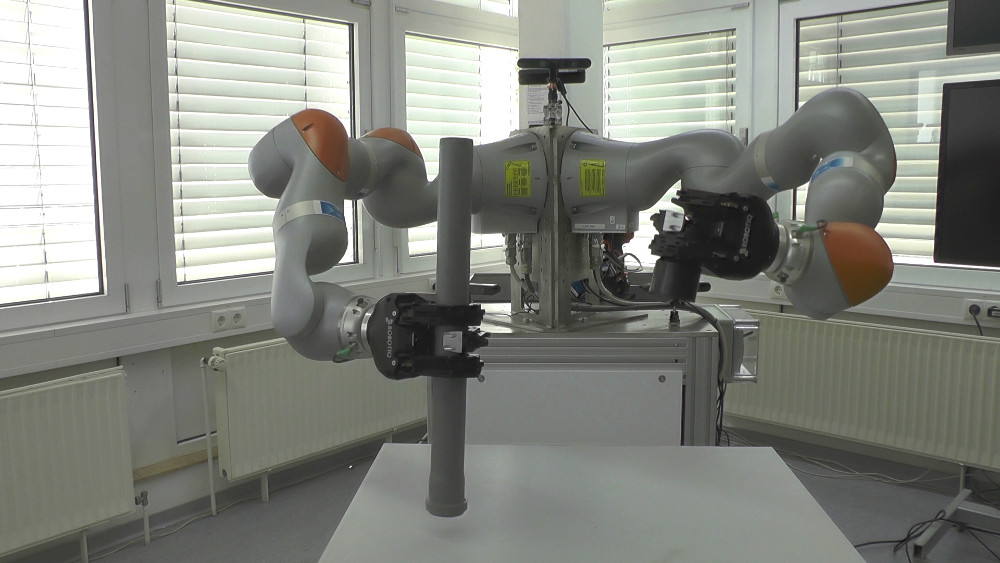}\vspace{0.1cm}\\
\includegraphics[width=0.4\linewidth]{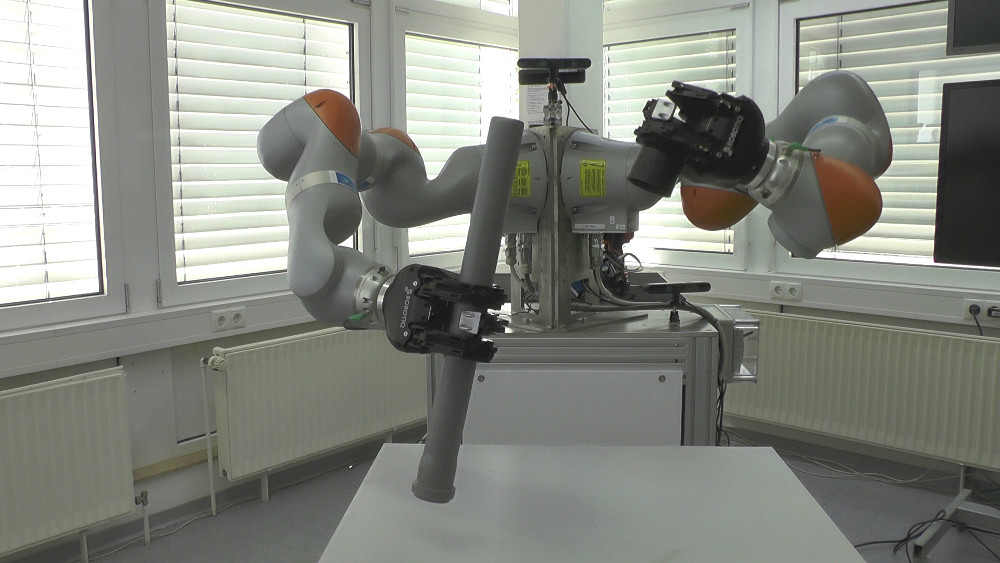}
\includegraphics[width=0.4\linewidth]{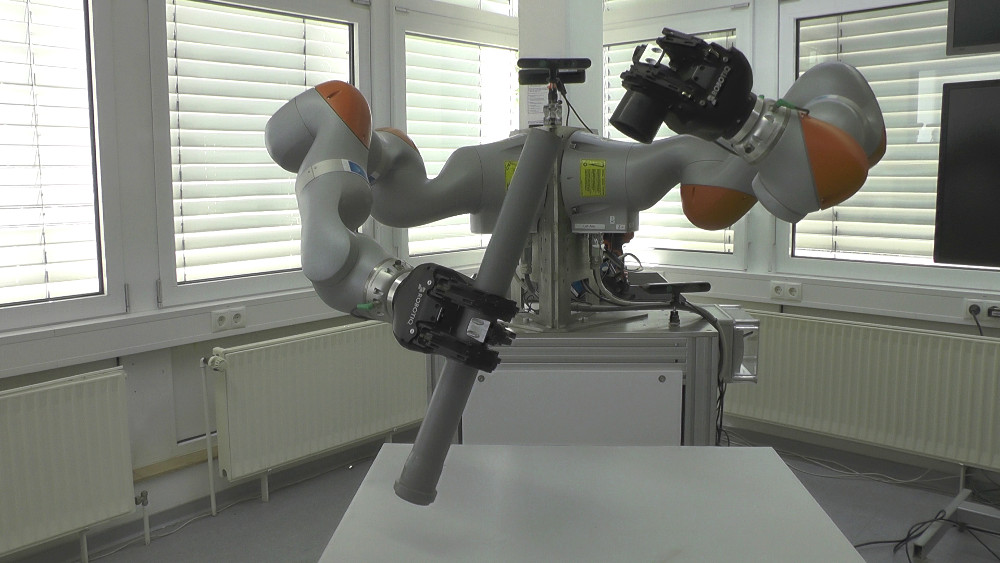}\vspace{0.1cm}\\
\includegraphics[width=0.4\linewidth]{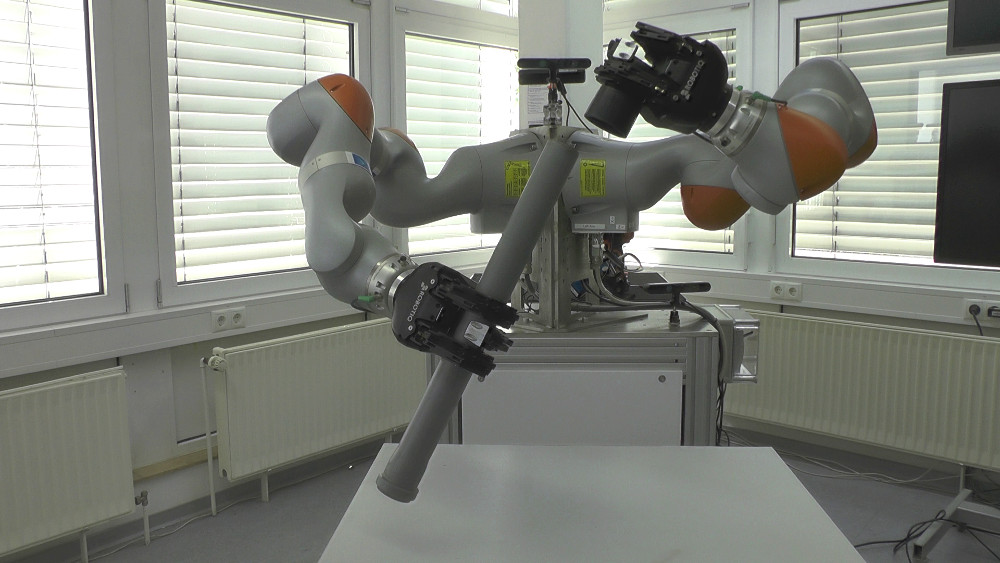}
\includegraphics[width=0.4\linewidth]{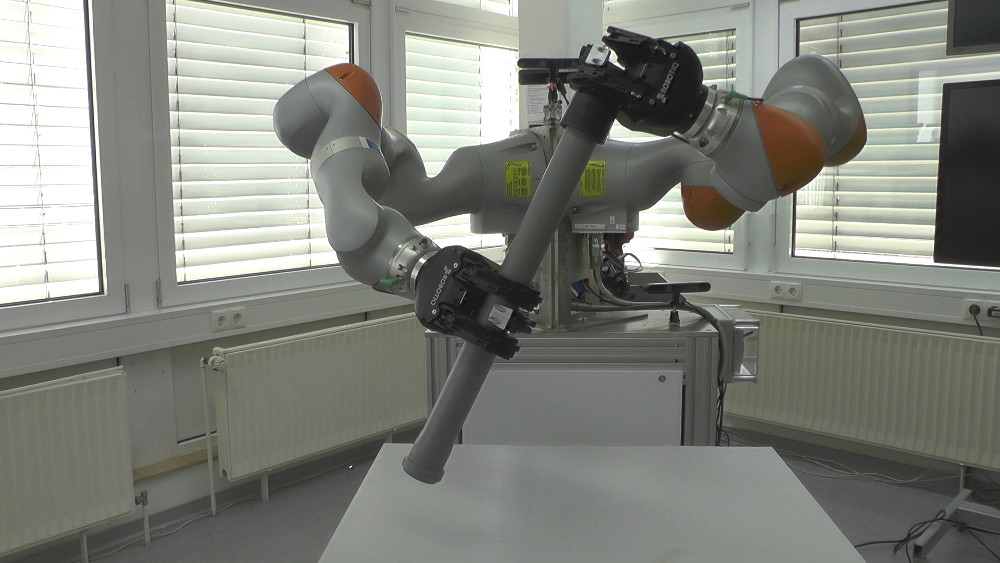}
\caption{Reproduction of the \textit{Assembly} task in context $A_{11}$}
\label{fig:results_assembly}
\end{figure}

\subsection{Reproduction of Task Constraints}

As described in section~\ref{sec:constraint_estimation}, a joint distribution $\mathcal{P}(\mathbf{v},\mathbf{x},\mathbf{\kappa})$  is learned using the recorded context and motion data $[\mathbf{X},\mathbf{V},\mathbf{\mathcal{K}}]$. We use a Dirichlet Process Gaussian Mixture Model to model the distribution.  For all tasks, we set the number of mixture components to $K=50$ and let the Dirichlet Process decide automatically on the effective number of mixtures. Reproduction of the task constraints is achieved as described in Algorithm~\ref{alg:reproduction}.

We evaluate the ability of the approach to generalize with respect to previously unseen situations, e.g., a new start pose $\mathbf{x}_0$ or category of the task. The latter is thereby described by the context vector $\mathbf{\kappa}$. The results are displayed in Figure~\ref{fig:results_motion_constraints} and explained in the following.

\subsubsection{Rotate Object}  The model is trained for clockwise rotation direction using both arms with data from the contexts $\{R_{11},R_{13},R_{15}\}$. Figure~\ref{fig:results_rotate_panel_varying_panel_size} shows the reproduction in the test contexts $\{R_{12}, R_{14}\}$, which represent previously unseen object sizes. As it can been seen the trained model is able to generalize over the size of the manipulated object. Figure~\ref{fig:results_rotate_panel} shows video snapshots of the reproduction of the \textit{Rotate Object} task in context $R_{12}$. 

Next, we train the model using clockwise rotation with both arms and anticlockwise rotation using only a single arm. We evaluate the learned model using \textit{anticlockwise} rotation using \textit{both arms}, a previously unseen context. Thus, we use the contexts $\{R_{11} \ldots R_{15}, R_{31},R_{32},R_{41},R_{42}\}$ for training and contexts $\{R_{21}, R_{23}, R_{25}\}$ for evaluation. The results are displayed in Figure~\ref{fig:results_rotate_panel_varying_rot_dir}. As can be seen here, the approach is able to generalize with respect to a change of the rotation direction (using both hands). While the model was trained with single arm motions for a counterclockwise rotation direction, it is able to generate dual-arm motions with both arms in the same rotation direction. 

The reproduction of this task is also illustrated in the accompanying video \href{anc/04_reproduction_rotate_panel.mp4}{04\_reproduction\_rotate\_panel.mp4}.

\subsubsection{Collaboration} Here, we train the model using $D=6$ of the  demonstrations for the fixed context $C_{11}$, which have varying start poses. We use the remaining $D=4$ demonstrations with unknown start poses for evaluation. The results in Figure~\ref{fig:results_carry_load_varying_start_pos} show the capability of the approach to generalize about different start poses. For the sake of clarity only the $xyz$-position is illustrated.  Figure~\ref{fig:results_carry_load} shows video snapshots of the \textit{Collaboration} task in context $C_{11}$. The results are also illustrated in the accompanying video \href{anc/05_reproduction_collaboration_a.mp4}{05\_reproduction\_collaboration\_a.mp4}.

\subsubsection{Assembly} Finally, we train the model using $D=6$ demonstrations from the assembly task with varying start poses (fixed context $A_{11}$) and use the remaining $D=4$ demonstrations with previously unknown start poses for evaluation. Figure~\ref{fig:results_assembly_varying_start_pos} shows the result. For the sake of clarity again only the $xyz$-position is shown. The results underline the ability of the model to generalize with respect to previously unknown start poses. Figure~\ref{fig:results_assembly} shows video snapshots of the \textit{Assembly} task in context $A_{11}$.  The results are also illustrated in the accompanying video \href{anc/06_reproduction_assembly.mp4}{06\_reproduction\_assembly.mp4}.

\begin{figure}
\centering
\begin{subfigure}{0.48\linewidth}
\includegraphics[width=\columnwidth]{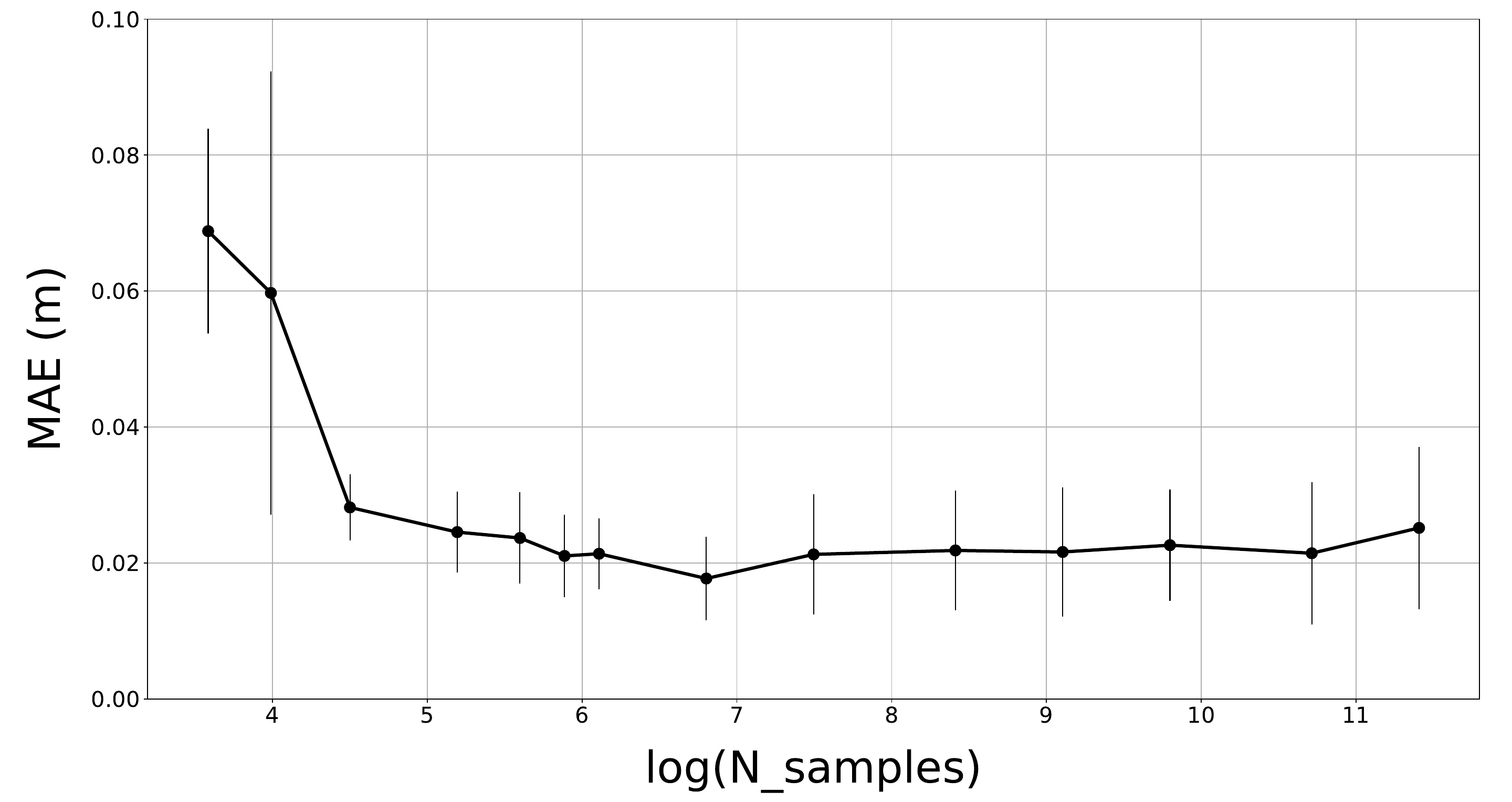}
\caption{Learning curve GMR}
\label{fig:learning_curve_gmr}
\end{subfigure}
\hfill
\begin{subfigure}{0.48\linewidth}
\includegraphics[width=\columnwidth]{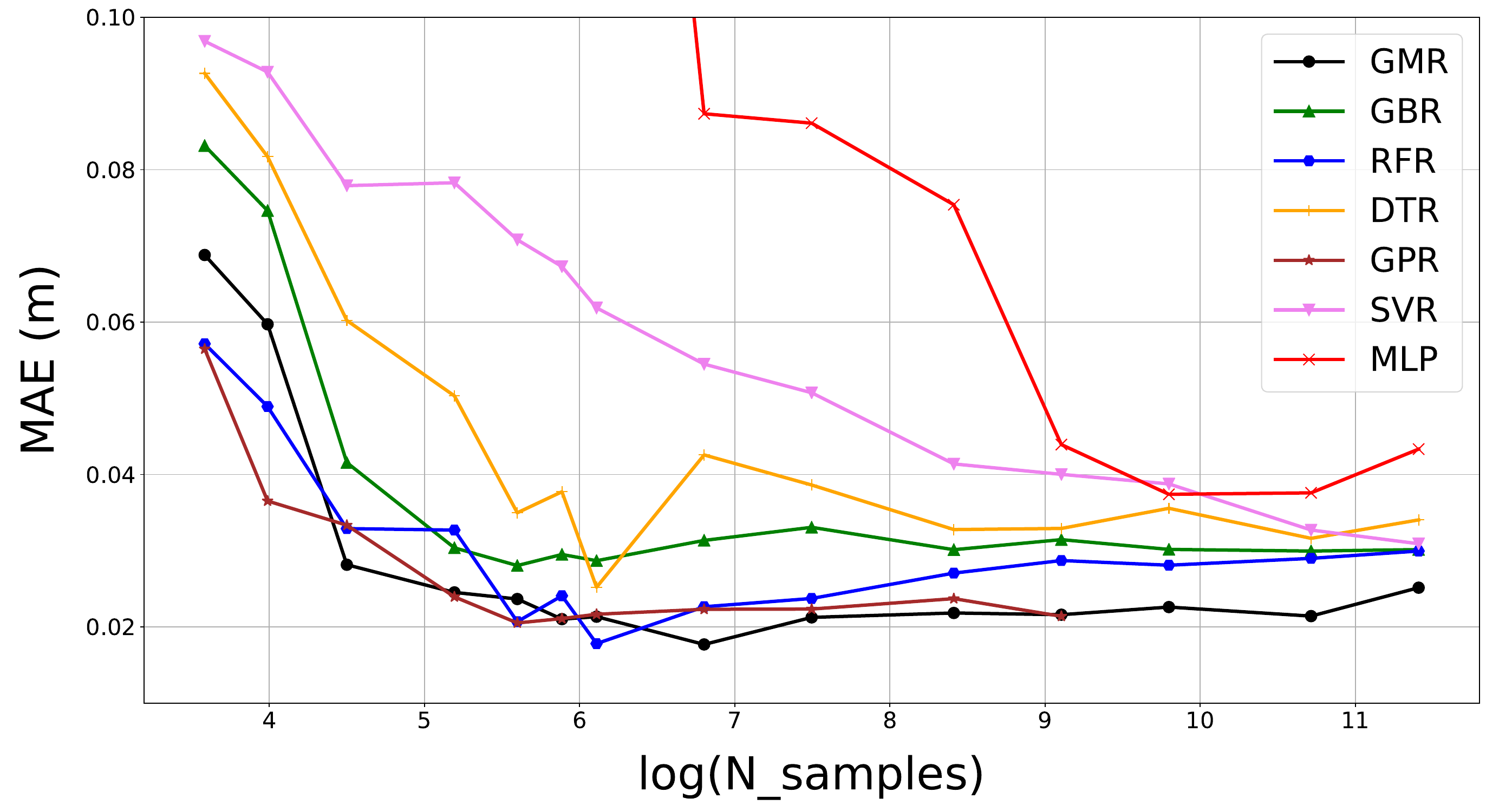}
\caption{Comparing different regressors\footnotemark}
\label{fig:learning_curve_all}
\end{subfigure}
\caption{Learning curves for the \textit{Rotate Object} task,  x-axis: Number of training samples, y-axis: mean absolute error between mean of the demonstrations and predictions made by the model for the contexts $R_{11} \ldots R_{15}$ (clockwise rotation) and $R_{21} \ldots R_{25}$ (anti-clockwise rotation) }
\label{fig:results_rotate_panel_2D}
\end{figure}
\footnotetext{GMR - Gaussian Mixture Regression, GBR - Gradient Boosting Regression, RFR - Random Forest Regression, DTR - Decision Tree Regression, GPR - Gaussian Process Regression, SVR - Support Vector Regression, MLP - Multi-layer Perceptron, All implementations are taken from scikit-learn~\cite{sklearn_api}}

\subsubsection{Model Performance}
\label{sec:model_performance}

We analyze the performance of our approach using the contexts $(R_{11} \ldots R_{15})$ and $(R_{21} \ldots R_{25})$ of the \textit{Rotate Object} task (see Table~\ref{tab:contexts_rotate_object}). The model is trained with $C-1$ of these contexts using grid search and leave-one-out cross validation for hyper-parameter tuning. Then we evaluate the resulting model by measuring the error between the mean of the demonstrations and the reproduced motion (predictions made by the model) for the remaining (unknown) context. Every context is used once for evaluation, so we perform $C$ evaluations in total. As a performance measure we use the mean absolute error (MAE) over all evaluations. 

 Figure~\ref{fig:learning_curve_gmr} shows the learning curve (Number of training samples vs. MAE including a single standard deviation depicted as error bars) for the \textit{rotate object} task. It can be seen that the motion can be reproduced reliably in unknown contexts with relatively low error (approx. $2cm$ on average). In Figure~\ref{fig:learning_curve_all} we compare the GMR learning curve with other regression methods user for motion synthesis. It can be seen that GMR requires a relatively low number of training samples to achieve good generalization capabilities (low reproduction error in unknown contexts) and provides a low overall reproduction error in general.

\begin{figure}
    \centering
    	\begin{subfigure}{\textwidth}
        \centering
        \includegraphics[width=0.49\linewidth]{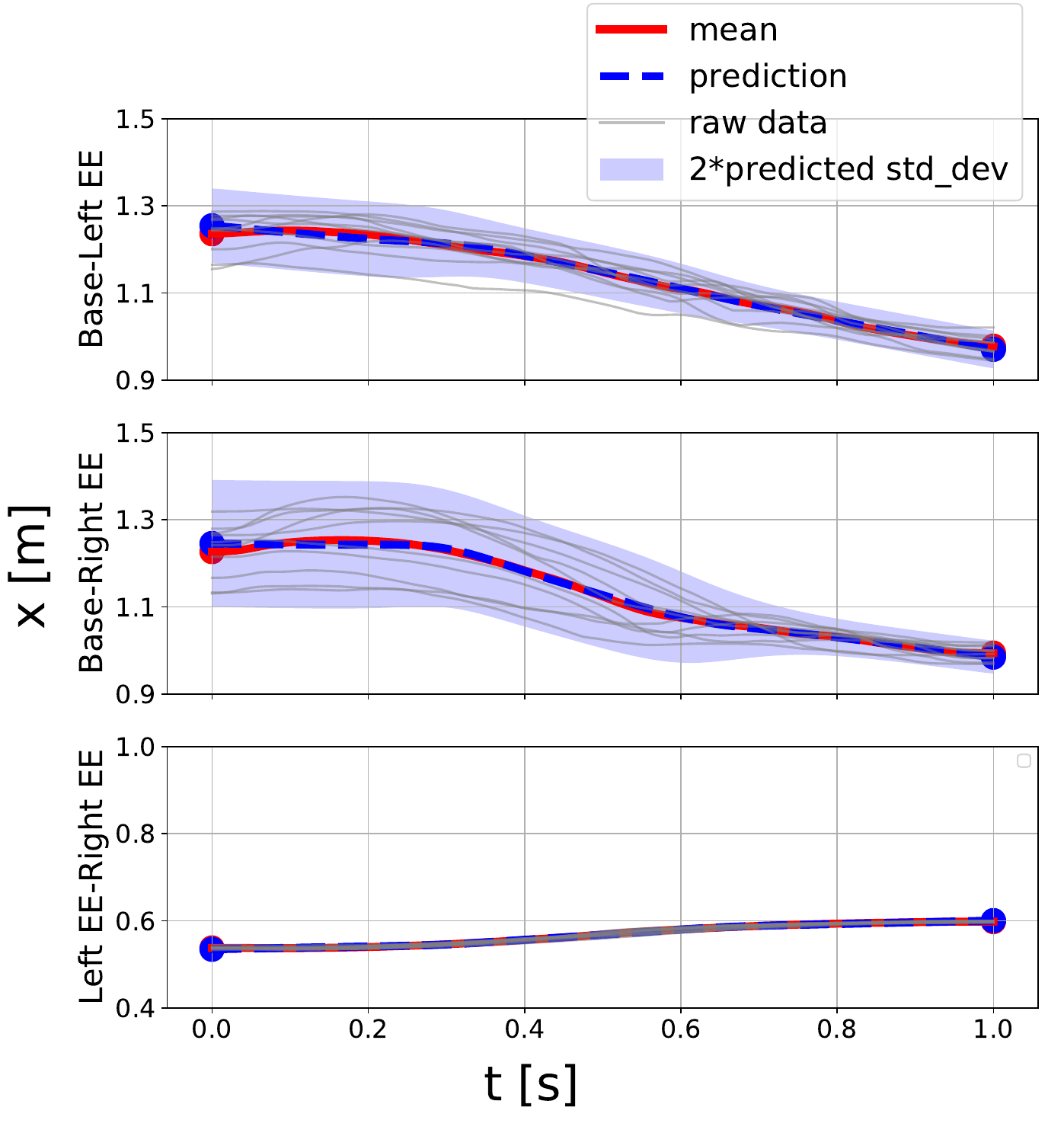}
        \includegraphics[width=0.49\linewidth]{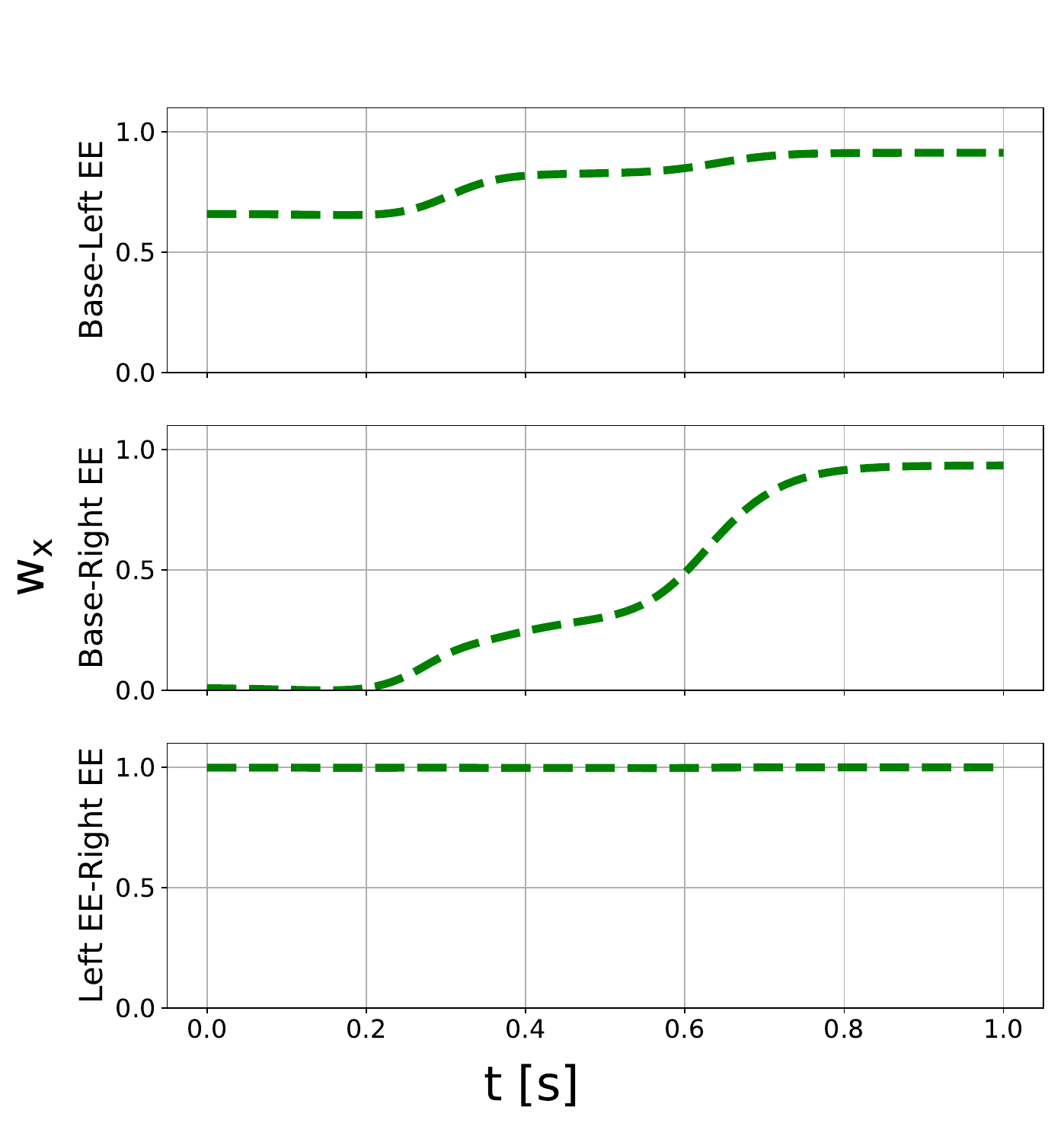}
        \caption{\emph{Rotate Object} task (only x-position, fixed context $R_{15}$). }
        \label{fig:results_rotate_panel_task_weight_estimation}	
    	\end{subfigure}\\ \vspace{0.2cm} 
    \centering
    	\begin{subfigure}{\textwidth}
        \centering
\includegraphics[width=0.49\linewidth]{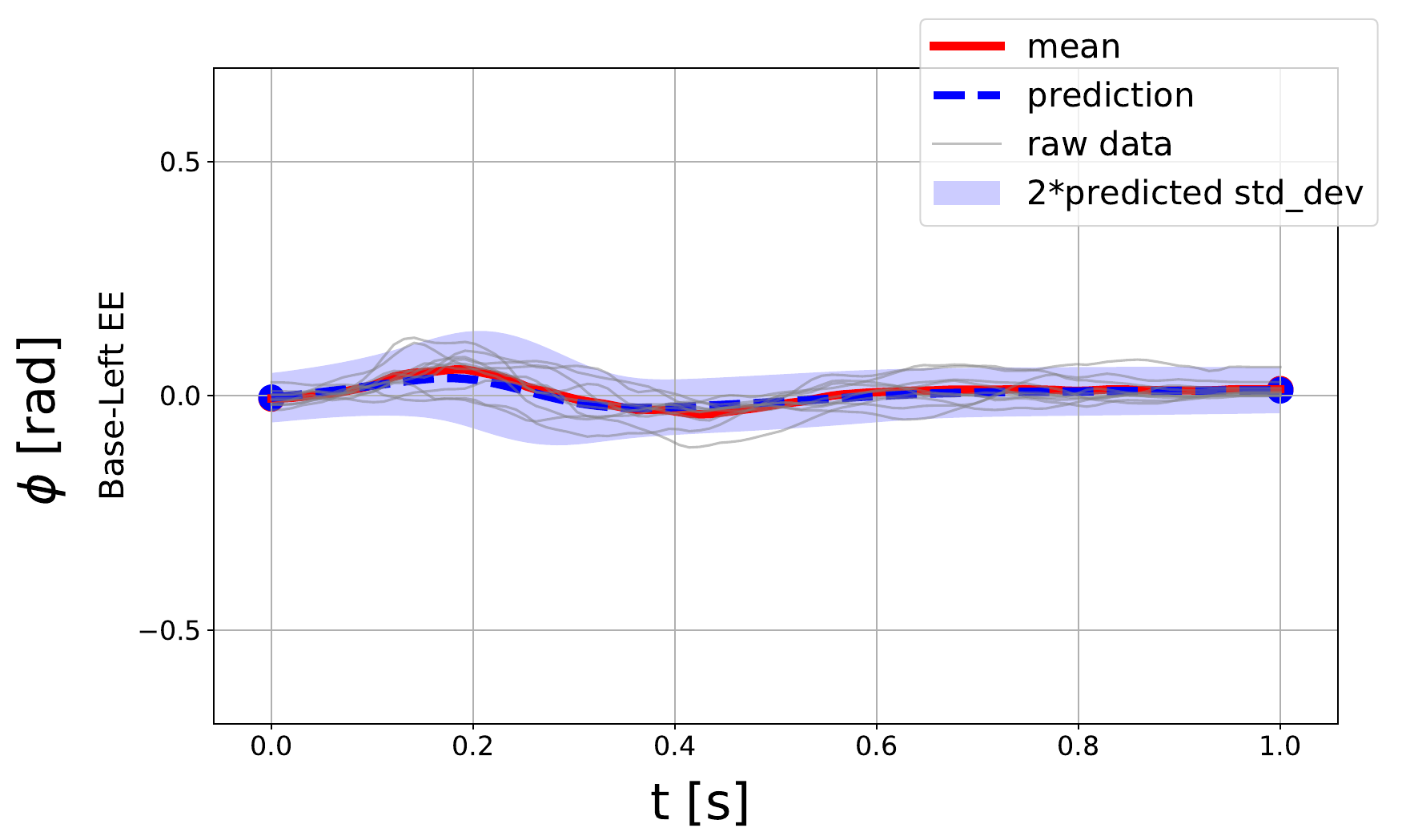}
\includegraphics[width=0.49\linewidth]{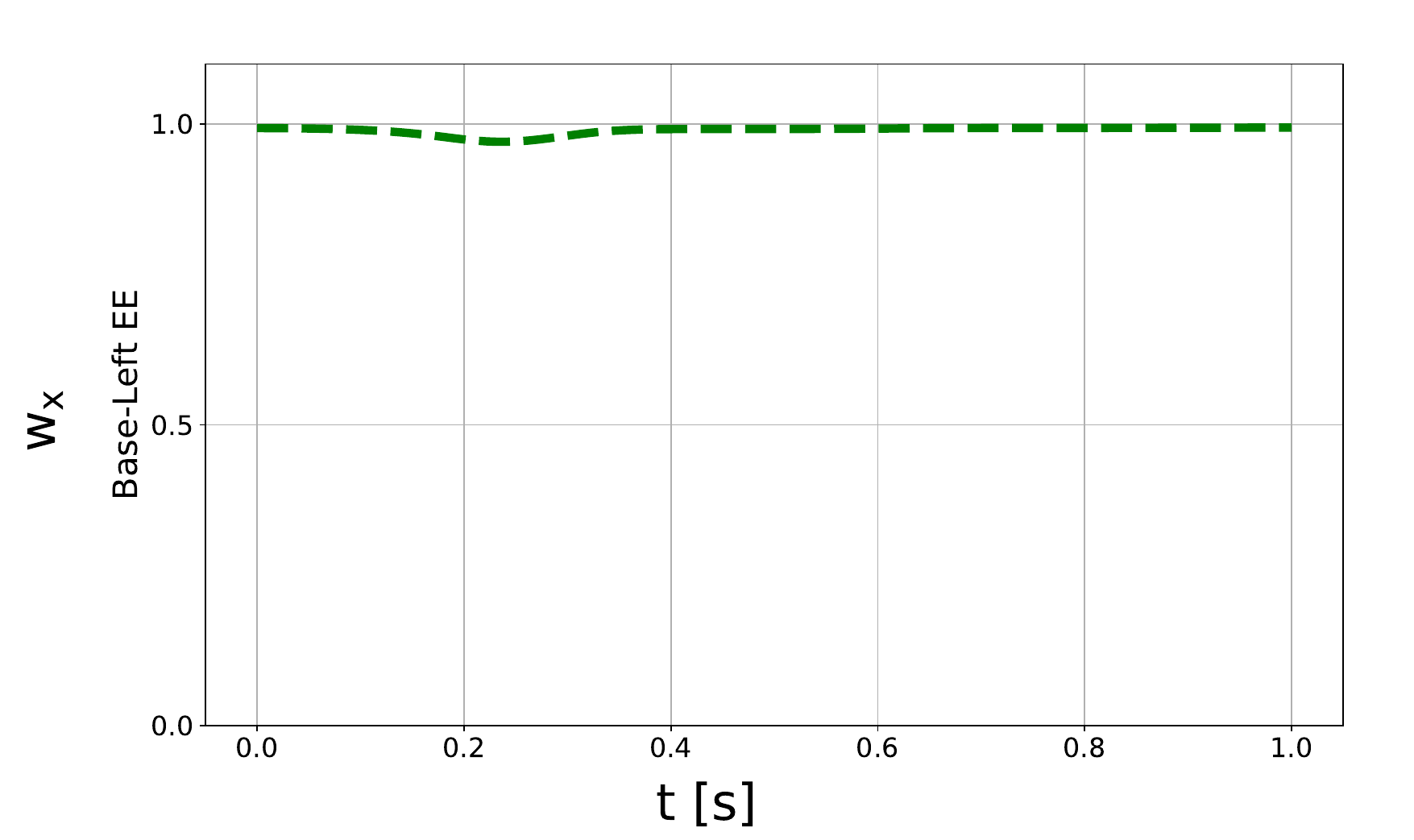}
        \caption{\textit{Collaboration} task (only $\phi$-orientation),  Reproduction in context $C_{11}$: Without tilting}
        \label{fig:results_carry_load_task_weight_estimation_a}
    	\end{subfigure} \\ \vspace{0.2cm}
    	\begin{subfigure}{\textwidth}
        \centering
\includegraphics[width=0.48\linewidth]{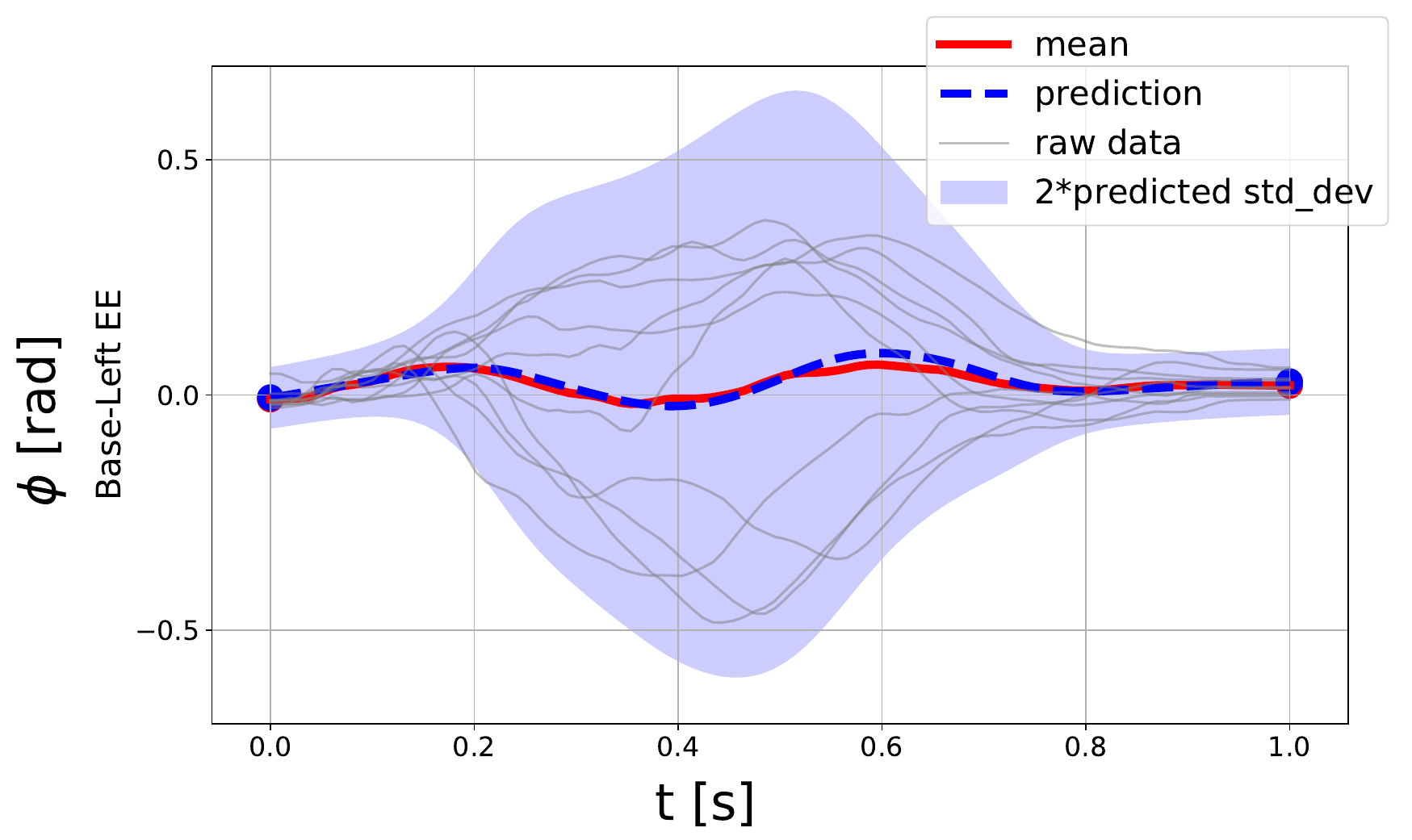}
\includegraphics[width=0.48\linewidth]{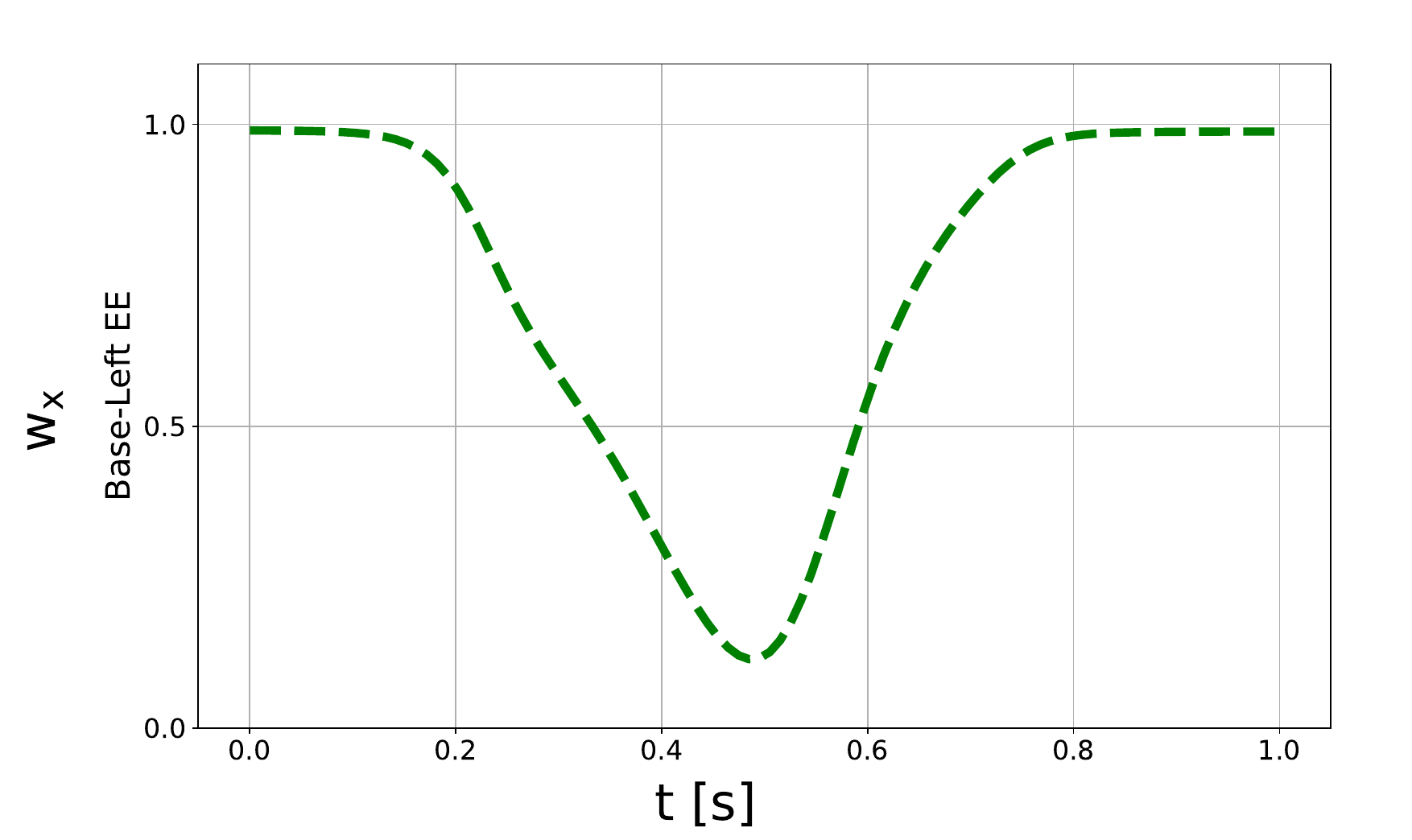}
        \caption{\textit{Collaboration} task  (only $\phi$-orientation), Reproduction in context $C_{12}$: With tilting}
        \label{fig:results_carry_load_task_weight_estimation_b}
    	\end{subfigure}
    \caption{Estimation of task weights: Temporal, inter-constraint and context adaptation. \textit{Left}: Reproduction of task constraints and variance, \textit{Right}: Estimation of task weights according to (\ref{eq:weight_estimate}).}
\end{figure}

\subsection{Estimation of Soft Task Priorities}

Next we evaluate the capability of the approach to estimate suitable task priorities and adapt them (a) over time (during task execution), (b) with respect to different constraint variables and (c) with respect to different contexts.  As described in section~\ref{sec:learning_task_constraints}, we estimate the task priorities from the variance in the user demonstrations according to Equation~(\ref{eq:weight_estimate}). A large variance corresponds to small task weights and vice versa. 

\subsubsection{Temporal and Inter-Constraint Adaptation}
Figure~\ref{fig:results_rotate_panel_task_weight_estimation} shows the reproduction of the \textit{Rotate Object} task (only x-axis), along with the demonstrated motions, the mean of the demonstrations and the estimated confidence interval $2\sigma$. Since we chose different start poses, the motion initially shows a large variance for the constraints \textit{Base-Left EE} and \textit{Base-Right EE} and becomes smaller during task execution, since we try to bring the object to the same final pose in each demonstration. Accordingly, the respective task weights are low in the beginning and increase during the course of the task. In contrast to that, the constraint \textit{Left EE - Right EE} has a low variance and, accordingly, a large task weight during the whole task. This is because the relative motion of the grippers is constrained by the object that they are holding. The results show that the estimated (soft) task priorities reflect the importance of the different constraints. In this case this means that the relative pose of the end effectors is more important than the pose of each individual end effector. Furthermore, it shows that the prioritization of constraints can be adapted during the course of a given task. 

\subsubsection{Context Adaptation}
Figure~\ref{fig:results_carry_load_task_weight_estimation_a} shows the effect of estimating the task weights in different contexts for the \textit{Collaboration} task (only $\phi$-rotation). We estimate $\mathbf{\hat{v}}_t, \mathbf{\hat{x}}_t$ as described in section~\ref{sec:constraint_estimation} for context $C_{11}$ (Figure~\ref{fig:results_carry_load_task_weight_estimation_a}), as well as for context $C_{12}$ (Figure~\ref{fig:results_carry_load_task_weight_estimation_b}) and compare the resulting task weights. Since we allow tilting the load during the motion in context $C_{12}$, the variance is large and the corresponding task weight drops during task execution. Compared to that the task weight in context $C_{11}$ remains high during the whole motion. The results are  illustrated in the accompanying video \href{anc/05_reproduction_collaboration_b.mp4}{05\_reproduction\_collaboration\_b.mp4}.

\begin{figure}[t]
\centering
\includegraphics[width=\columnwidth]{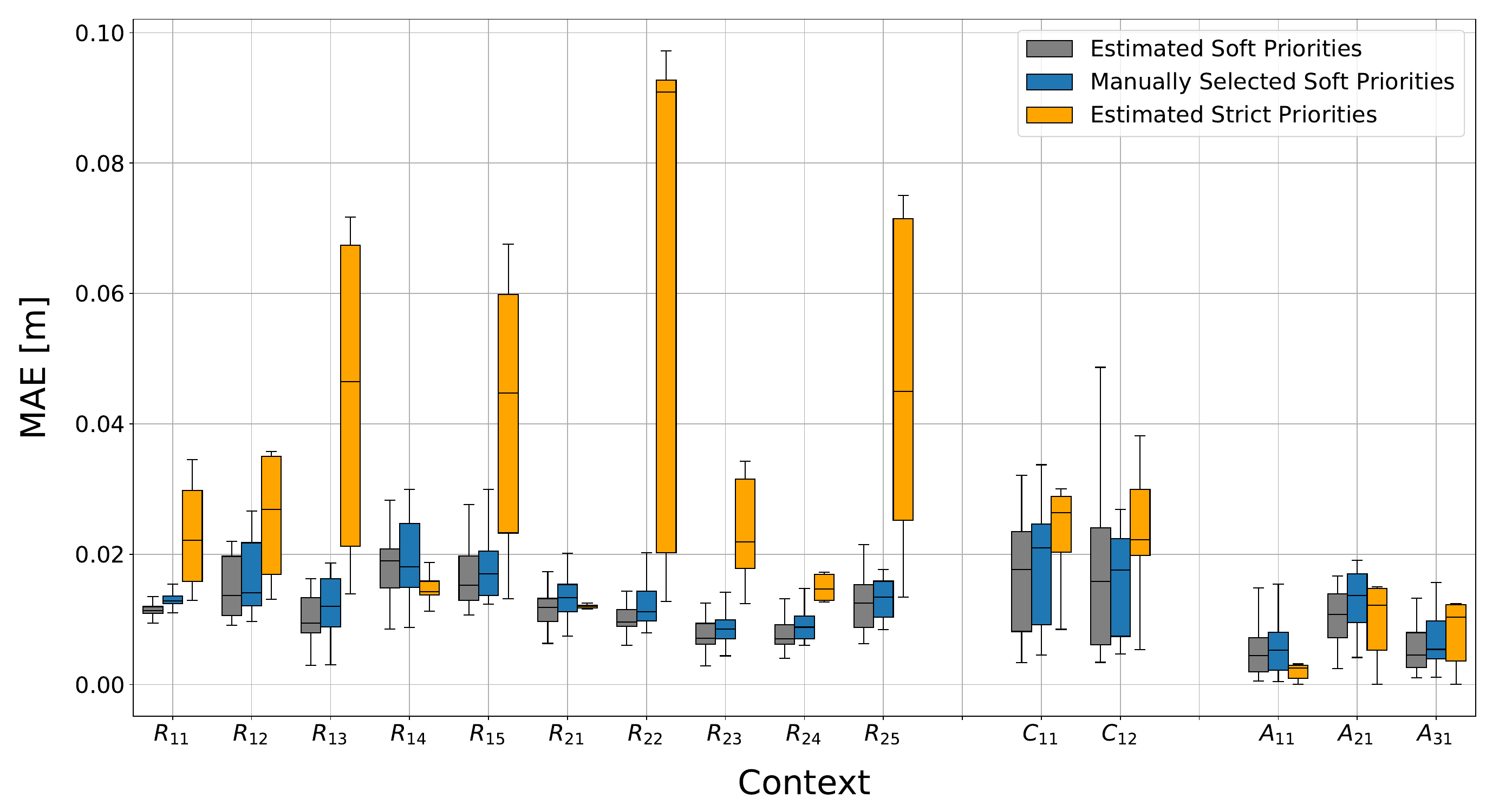}
\caption{Comparison of the reproduction error using three different methods for task prioritization: Estimated soft task priorities (our approach, grey),  Manually selected soft task priorities (blue) and Estimated strict hierarchies according to~\cite{Silverio2017} (orange).}
\label{fig:task_weight_comparison}
\end{figure}

\begin{figure}[t]
\centering
\begin{subfigure}{0.47\linewidth}
\includegraphics[width=\columnwidth]{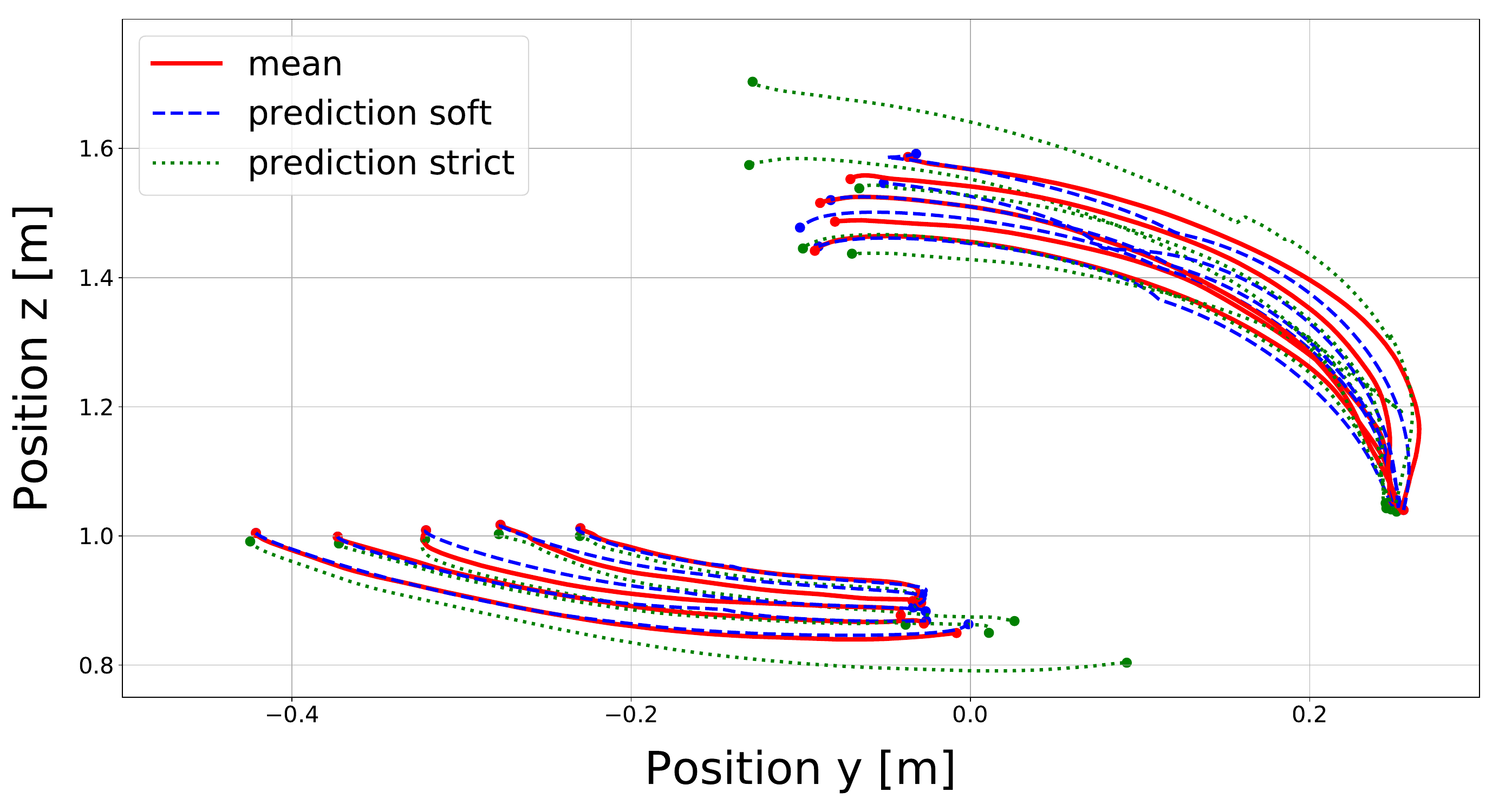}
\caption{\textit{Rotate Object} (clockwise, contexts $R_{11} \ldots R_{15}$)}
\end{subfigure}
\hfill
\begin{subfigure}{0.47\linewidth}
\includegraphics[width=\columnwidth]{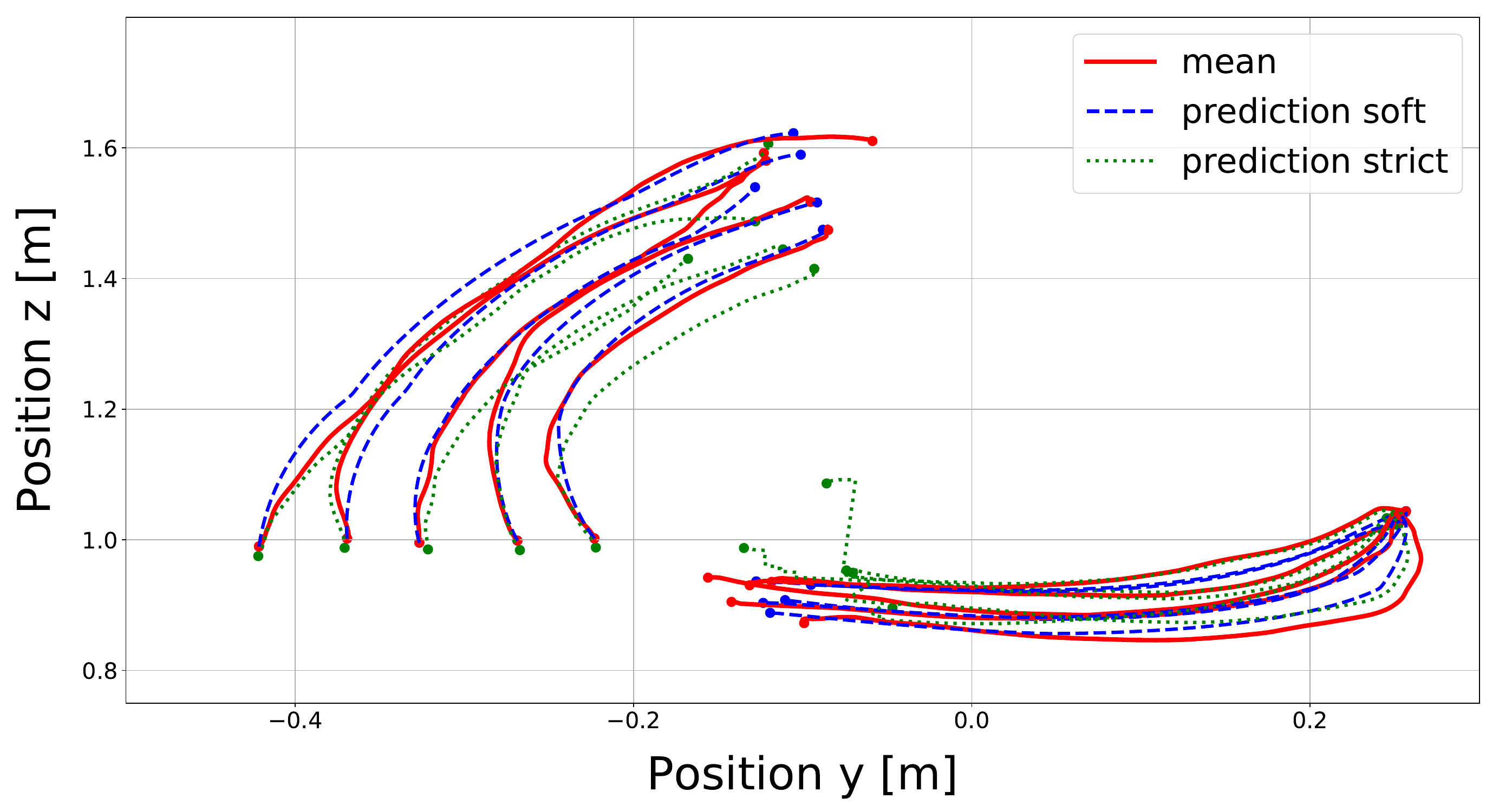}
\caption{\textit{Rotate Object} (anti-clockwise, contexts $R_{21} \ldots R_{25}$)}
\end{subfigure}
\vspace{0.3cm} \\ 
\begin{subfigure}{0.47\linewidth}
\includegraphics[width=\columnwidth]{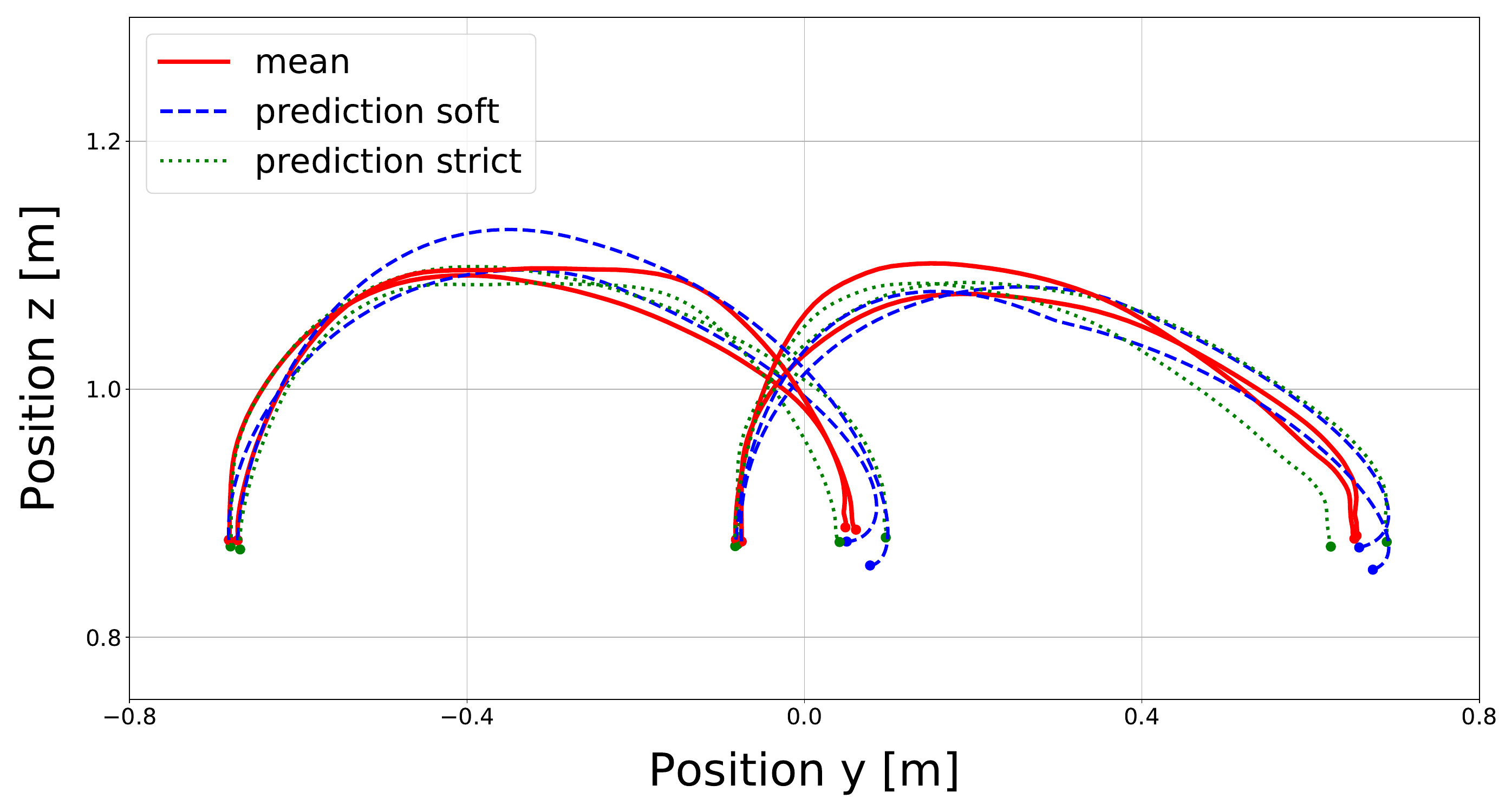}
\caption{\textit{Collaboration} (dual-arm, contexts $C_{11}, C_{12}$)}
\end{subfigure}
\hfill
\begin{subfigure}{0.47\linewidth}
\includegraphics[width=\columnwidth]{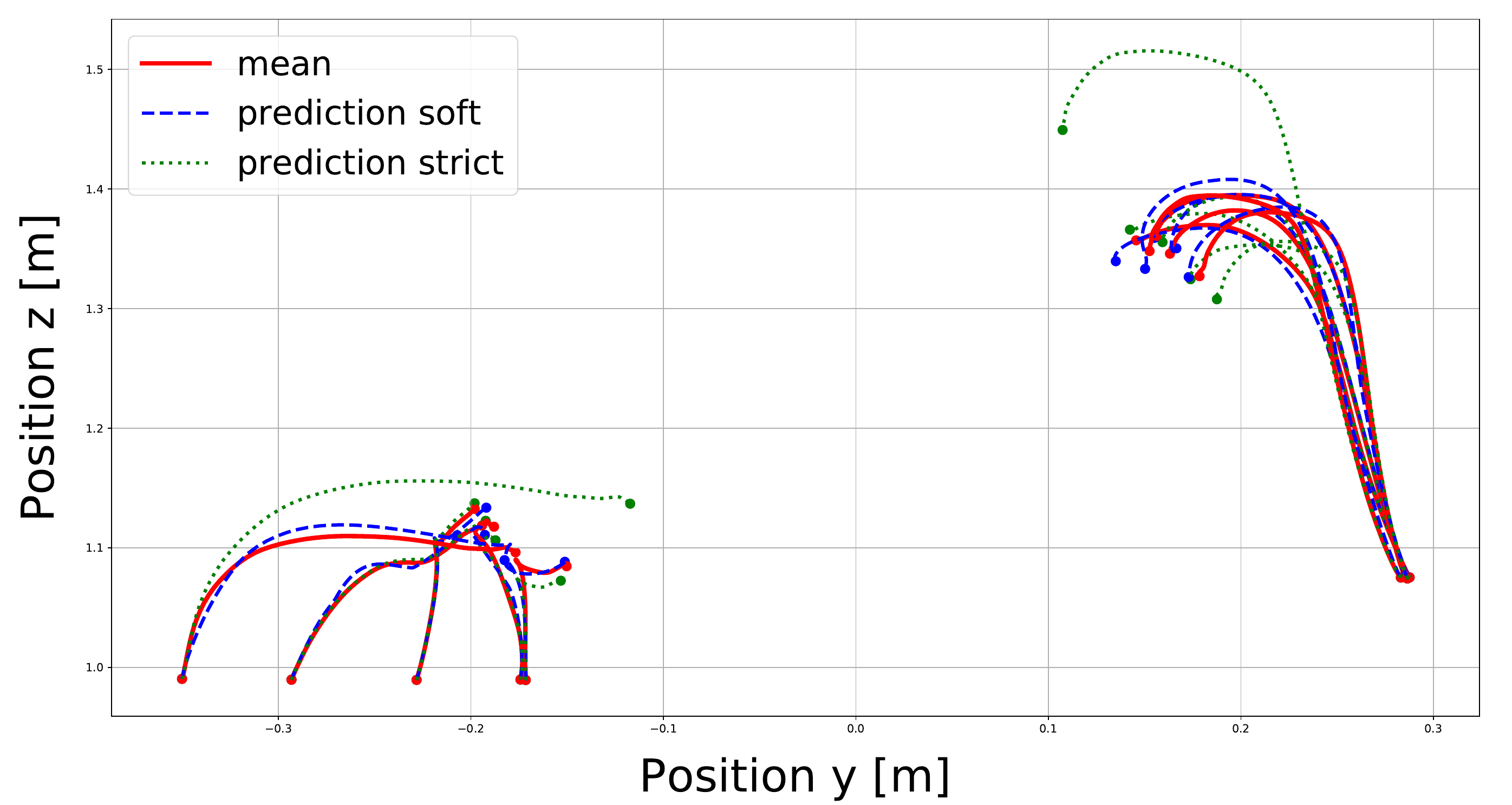}
\caption{\textit{Assembly} (context $A_{11}$, varying start poses)}
\end{subfigure}
\caption{Comparison of the reproduced motion using different methods for task prioritization: Estimated soft task priorities (our approach, blue dashed) and estimated strict task priorities according to~\cite{Silverio2017} (green dotted). The mean of user demonstrations is illustrated as solid red line.}
\label{fig:results_compare_strict_soft}
\end{figure}

\subsubsection{Comparison of Different Approaches for Task Prioritization}
\label{sec:results_prio_compare}

Finally, we compare our approach for estimating soft task priorities with two other methods for task prioritization: (1) Manually selected soft task priorities. Here we select a fixed value $w=1$ for all task weights during the complete motion. (2) Strict hierarchy estimation from data as described in~\cite{Silverio2017}. Here, strict prioritization is enforced through consecutive null space projections of the respective task Jacobians. The task hierarchy is thereby estimated from the variance in the user demonstrations, given a set of candidate hierarchies. The idea is that hierarchies with a low relevance produce a higher variability during demonstration than hierarchies with high relevance. For movement synthesis the candidate joint space velocities generated by each hierarchy are fused using a soft weighting scheme, where the a high variance in the user demonstrations corresponds to a low weight and vice versa. 

In our case, we have three tasks, denoted as \emph{Base}-\emph{Left EE} (\emph{left}), \emph{Base}-\emph{Right EE} (\emph{right}) and \emph{Left EE}-\emph{Right EE} (\emph{relative}). For evaluation, we choose the following candidate hierarchies:

\begin{center}
    \footnotesize
	 \begin{tabular}{ccccc}
  \midrule
     priority & highest & medium & lowest \\
   \midrule 
   $h_{1}$ & \emph{left}& \emph{right} & \emph{relative} \\
   $h_{2}$ & \emph{left} & \emph{relative}  & \emph{right}   \\
   $h_{3}$ & \emph{right} & \emph{left} & \emph{relative}   \\
   $h_{4}$ & \emph{right} &\emph{relative} & \emph{left}   \\
   $h_{5}$ & \emph{relative} & \emph{right} & \emph{left}   \\
   $h_{6}$ & \emph{relative} & \emph{left} & \emph{right}   \\
  \midrule
 \end{tabular}
\end{center}

where a lower prioritized task is executed in the null space of the higher prioritized one and is not disturbing the execution of the latter. 

The evaluation results are shown in Figure~\ref{fig:task_weight_comparison}. 
As in section~\ref{sec:model_performance}, we measure the MAE between the mean of the user demonstrations and the predicted motion in multiple contexts\footnote{see Table~\ref{tab:contexts} for an overview over the contexts} using different approaches for task prioritization. Figure~\ref{fig:results_compare_strict_soft} shows the resulting motions (only yz-positions). 

From these figures we can derive the following results: (1) Our approach for estimating soft task priorities results in a lower reproduction error compared to the use of fixed task weights. Apart from that, it allows a bigger flexibility for executing additional tasks like e.g., collision avoidance. (2) Our approach results in a lower reproduction error when comparing to the method described in~\cite{Silverio2017} (fusion of strict hierarchies). This is due to the following reasons: In all three evaluation tasks, we have an over-constrained case (18 constraints, but only 14 degrees of freedom). Obviously, strict prioritization is not useful in such a case since it results in a bad tracking performance for the tasks with lowest priority. Fusing the candidate hierarchies with a soft weighting scheme cannot overcome this issue, at least not with the training data that we acquired. The user demonstrations that we provide are not optimal for this approach, since we did not focus on explicitly demonstrating certain hierarchies. (3) The mean reproduction error of our method is in the magnitude of around $0.005m - 0.02m$. Since we are not dealing with high precision tasks here and the KUKA arms have integrated joint level compliance controllers that may compensate small inaccuracies, the resulting reproduction errors are acceptable. Further improvements can be achieved by obtaining more examples from demonstration.

\subsection{Discussion}\label{sec:discussion}

In the previous sections, we experimentally evaluated the approach for automatic derivation and contextual adaptation of task constraints. We found that the use of GMM-GMR offers an intuitive way to program robot tasks using constraint-based control approaches and derive suitable task priorities automatically from user demonstrations. Furthermore, the learned models can generalize to a certain degree with respect to context changes that reflect variations of the environment or the given task. As a result, we achieve a better performance with respect to comparable methods and the user does not need reprogram every novel situation, but can rely on the generalization capabilities of the model. On the long-term we strive towards a decision process, where task constraints can be described on a semantic level and their numerical counterparts are automatically selected depending on the current situation. Such a framework has the potential to greatly increase the usability and autonomy of robotic systems. 

Our approach as described in section~\ref{sec:constraint_estimation} relies on good quality user demonstrations that reflect the given task constraints. If the user demonstrations do not cover the constraint space well, the estimated task weights might be  suboptimal, e.g., the solution might be unnecessarily over- or under-constrained. For example, when teaching a robot to place an object on a table, its motion is obviously constrained in the direction perpendicular to the surface (z-direction), while it can place the object quite freely somewhere on the table (xy-direction). The user demonstrations should address this by varying the target position on the surface as much as possible. This results in a high variance in xy-direction and a low priority for the corresponding constraints according to~(\ref{eq:weight_estimate}). Conversely, the z-direction is assigned a low variance and a high priority. This way, the resulting task priorities will allow the robot to fulfill additional tasks in xy-direction like e.g obstacle avoidance. This simple example shows that a thorough design of the experiments for data acquisition is a crucial part of the approach. Active learning approaches can be useful here to extend the programming-by-demonstration paradigm and supply the user with hints on "what to teach next" (see e.g.,~\cite{Calinon2007a}).

The approach does not scale for many task frames since the number of resulting constraints is equal to the number of the possible combinations drawn from the set of task frames. The selection of task frames is a design choice by the user and requires expert knowledge on whether a frame is relevant or not. Eventually, the information whether a task frame introduces redundant information on the task could be derived from the data acquired in user demonstrations. Redundant or irrelevant task frames could then be ignored when training the model. 

In this work we decided to use categorical variables for representing the context to ease the labeling of the demonstrations for the user. Although Gaussians are usually not well suited to represent categorical variables, GMM's are able to fit the data quite well if suitable regularization of the model parameters is done. In future, a different representation of the categorical variables like binomial distributions could be chosen.

\section{CONCLUSION} \label{sec:conclusion}

The combination of constraint-based control and imitation learning has great potential.  While imitation learning offers an intuitive user interface to define new robot tasks, constraint-based task specification and control provides a powerful and flexible tool to compose complex robot behaviors. The seamless integration of both promises improvements in terms of usability, general applicability and autonomy of complex robotic systems with many dof. For examples, it is straightforward to integrate expert knowledge by manually programming some constraints, while learning others that cannot be easily specified. 

One shortcoming of our approach is that for each demonstration, the current context has to be labeled by the human expert. Thus, a logical next step would be to classify the current context from the recorded data and determine whether a demonstration belongs to a known or to an unknown context. Another issue is that the task frames have to be selected by the user in advance and, for a large number of task frames, the approach does not scale. Thus, it would also be useful to select optimal task frames from the user demonstrations, e.g., use frames that maximize the information gain. We plan to investigate both problems in future. Moreover, we would like to apply our approach to more complex scenarios including different types of constraints (contact forces, obstacles, ...) and more complex robots (e.g., humanoids). Finally, the estimated task weights from the model might not be optimal, since they strongly depend on the quality of  user demonstrations. For example, the computed task weights might unnecessarily over-constrain the system, leaving less dof for additional tasks. Thus, we would like to add an optimization step that improves the task weights with respect to a suitable criterion, like e.g., manipulability.


\section*{ACKNOWLEDGMENT}

This work has been supported by a grant of the German Federal Ministry for Economic Affairs and Energy (BMWi, grant number 50RA1701).


\section*{References}

\bibliography{ref}

\end{document}